\documentclass[10pt,twocolumn,letterpaper]{article}

\usepackage[pagenumbers]{wacv} 

\usepackage{amsmath,amsfonts,bm}

\def\eqref#1{equation~\ref{#1}}

\def\1{\bm{1}}

\def\vb{{\bm{b}}}

\def\vp{{\bm{p}}}

\def\vr{{\bm{r}}}

\def\vt{{\bm{t}}}

\def\vv{{\bm{v}}}
\def\vw{{\bm{w}}}
\def\vx{{\bm{x}}}

\def\mB{{\bm{B}}}

\def\mF{{\bm{F}}}

\def\mI{{\bm{I}}}

\def\mL{{\bm{L}}}

\def\mT{{\bm{T}}}

\def\mW{{\bm{W}}}

\DeclareMathAlphabet{\mathsfit}{\encodingdefault}{\sfdefault}{m}{sl}
\SetMathAlphabet{\mathsfit}{bold}{\encodingdefault}{\sfdefault}{bx}{n}
\newcommand{\tens}[1]{\bm{\mathsfit{#1}}}

\def\tI{{\tens{I}}}

\newcommand{\method}{\mbox{ProLIP}\xspace}
\newcommand{\methodvalfree}{\mbox{$\text{ProLIP}_\varnothing$}\xspace}
\definecolor{wacvblue}{rgb}{0.21,0.49,0.74}
\usepackage[pagebackref,breaklinks,colorlinks,allcolors=wacvblue]{hyperref}

\usepackage{xspace}
\usepackage{algorithm}
\usepackage{listings}
\usepackage{multirow}
\usepackage{graphicx} 
\usepackage{caption}
\usepackage{subcaption}
\usepackage{booktabs}
\usepackage{amssymb}
\usepackage{makecell}
\usepackage{soul}
\usepackage{adjustbox}
\usepackage[table, dvipsnames]{xcolor}
\usepackage{pifont}     
\definecolor{shadecolor}{rgb}{0.95,0.95,0.95}
\definecolor{darkgreen}{RGB}{0,128,0}
\usepackage{arydshln}
\usepackage[accsupp]{axessibility}

\usepackage{array}
\newcolumntype{R}[2]{%
    >{\adjustbox{angle=#1,lap=\width-(#2)}\bgroup}%
    l%
    <{\egroup}%
}
\newcommand*\rot{\multicolumn{1}{R{45}{1em}}}

\newcommand{\var}[1]{{\scriptsize\,$\pm$#1}}

\newcommand{\best}[1]{\textbf{#1}}
\newcommand{\second}[1]{\underline{#1}}

\newcommand{\tick}{\textcolor{darkergreen}{\checkmark}}
\newcommand{\xmark}{\textcolor{red}{\ding{55}}}

\definecolor{lavender}{HTML}{E8DFF5} 
\definecolor{blush}{HTML}{FCE1E4}    
\definecolor{peach}{HTML}{FCF4DD} 
\definecolor{mint}{HTML}{DDEDEA}   
\definecolor{skyblue}{HTML}{DAEAF6}
\definecolor{darkergreen}{RGB}{0, 100, 0}

\def\wacvPaperID{118} 
\def\confName{WACV}
\def\confYear{2026}

\title{CLIP's Visual Embedding Projector is a Few-shot Cornucopia}

\author{Mohammad Fahes${^1}$ \quad Tuan-Hung Vu$^{1,2}$ \quad  Andrei Bursuc$^{1,2}$ \quad Patrick Pérez$^{3}$ \quad  Raoul de Charette$^{1}$
{}\\
$^1$ Inria \quad \quad \quad  $^2$ Valeo.ai \quad \quad \quad $^3$ Kyutai\\
}

\begin{document}

\twocolumn[{
\renewcommand\twocolumn[1][]{#1}%
    \centering
    \vspace{-1em}
    \maketitle
        \vspace{-1em}
	\begin{minipage}{0.31\linewidth}
		\centering\scriptsize
		\includegraphics[width=1.0\linewidth]{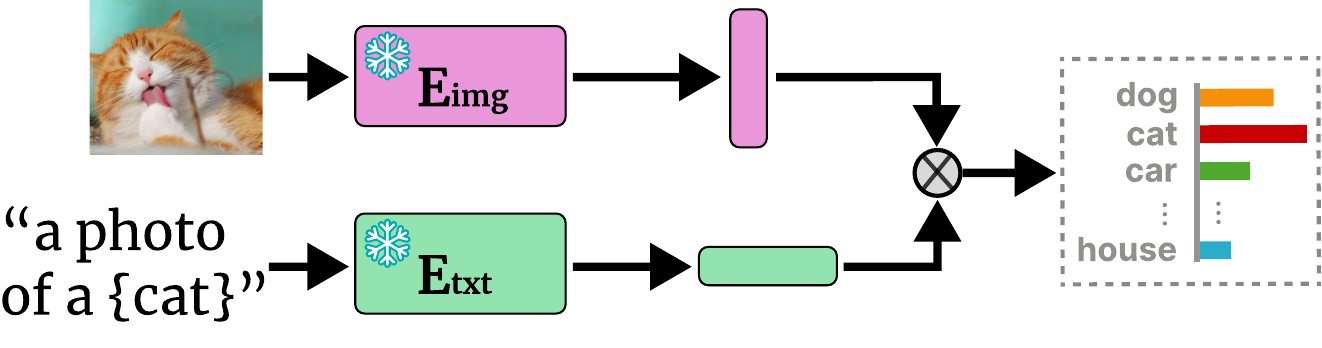}\\
		\textbf{(a) CLIP}
	\end{minipage}\hfill
	\begin{minipage}{0.28\linewidth}
		\centering\scriptsize
		\includegraphics[width=1.0\linewidth]{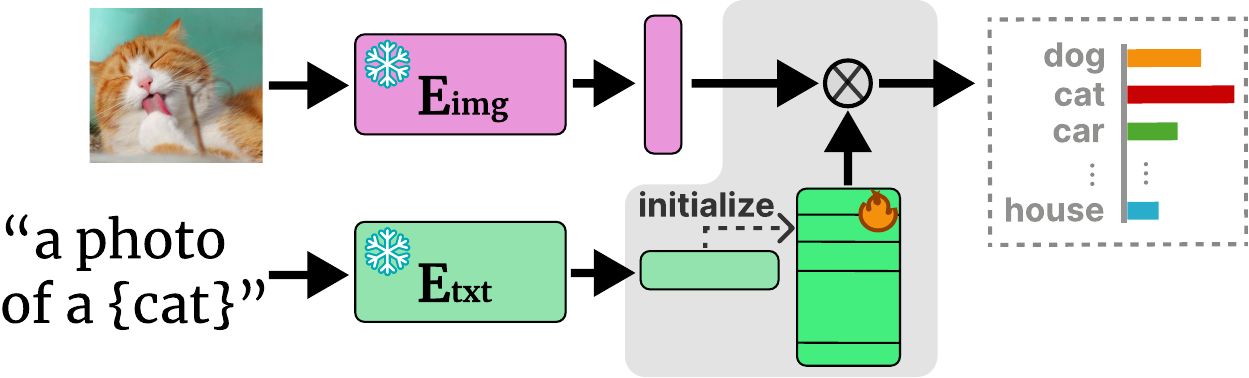}\\
		\textbf{(b) Linear Probing}
	\end{minipage}\hfill%
	\begin{minipage}{0.38\linewidth}
		\centering\scriptsize
		\includegraphics[width=1.0\linewidth]{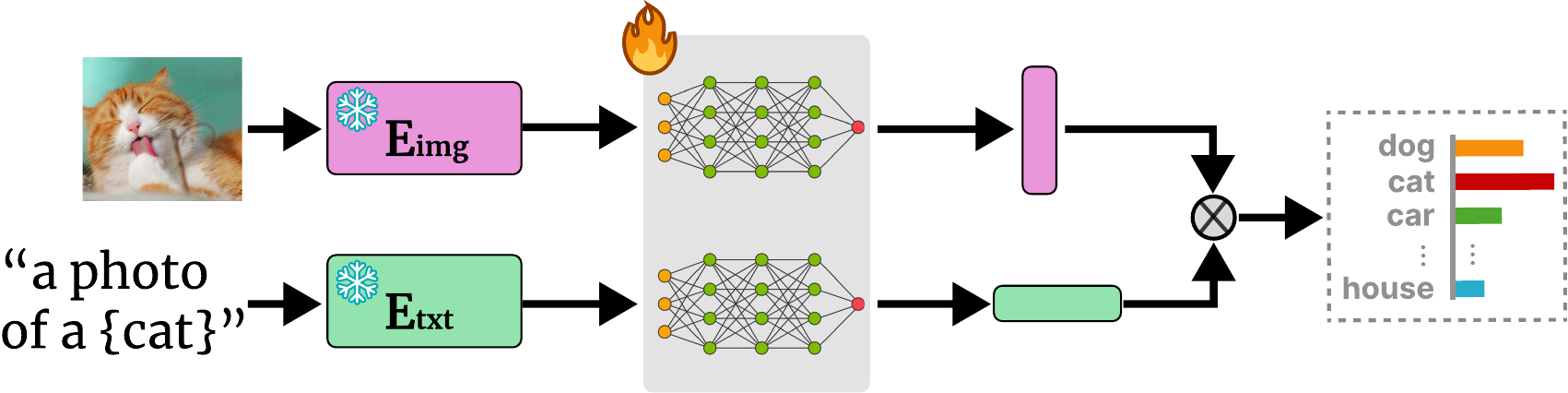}
		\textbf{(c) Adapters}
	\end{minipage}\\[0.5em]
	\begin{minipage}[t]{0.39\linewidth}
		\centering\scriptsize
		\includegraphics[width=1.0\linewidth]{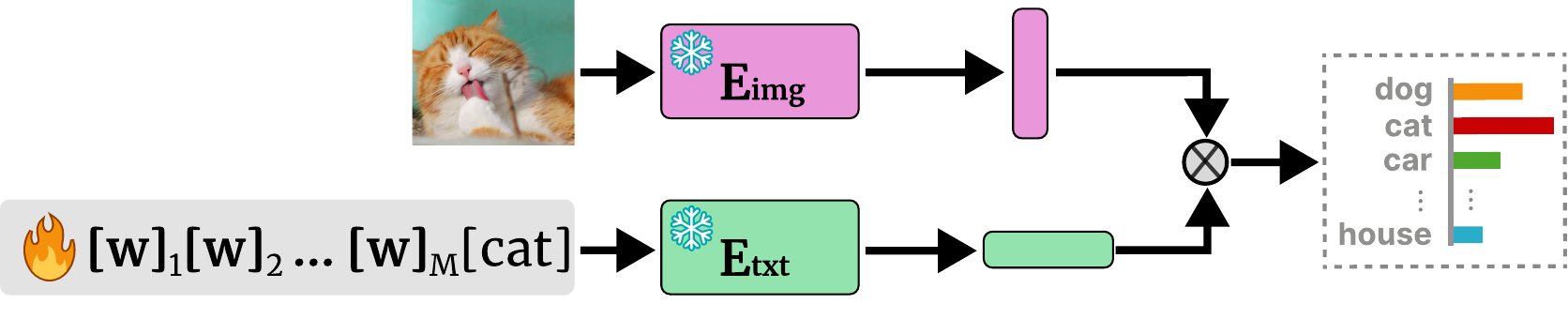}
		\textbf{(d) Prompt Tuning}
	\end{minipage}\hspace{0.1\linewidth}
	\begin{minipage}[t]{0.30\linewidth}
		\centering\scriptsize
		\includegraphics[width=1.0\linewidth]{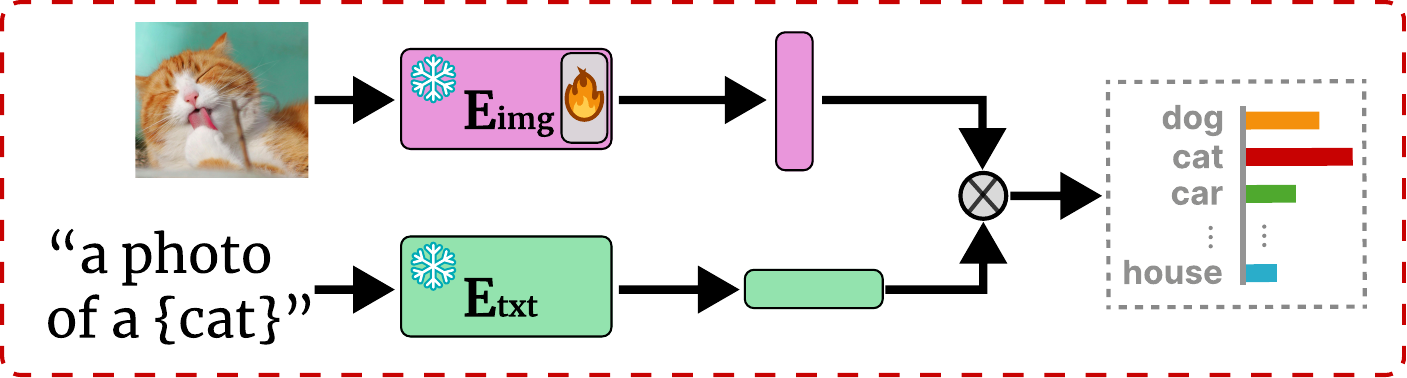}\\
		\textbf{(e) \method} (Ours)
	\end{minipage}\\
    \captionof{figure}{\textbf{Few-shot classification with CLIP.} \textbf{(a)} Using a pre-trained CLIP, \textit{zero-shot} classification is performed by measuring text and visual embeddings similarity. Among \textit{few-shot} adaptation strategies of CLIP, \textbf{(b)} Linear Probing~\cite{huang2024lp++,silva2024closer} trains a linear classifier of the visual features, \textbf{(c)} Adapters add external learnable MLPs~\cite{gao2024clip,zhang2022tip}, \textbf{(d)} Prompt Tuning learns word embeddings~\cite{zhou2022learning,zhou2022conditional,zhu2023prompt,chen2023plot}. Alternatively, we propose \textbf{(e)}~\method which does not introduce new weights
    and only fine-tunes the visual embedding linear projector.
    \\}
    \label{fig:teaser_figure}
}]%

\begin{abstract}
\vspace{-0.13em}
We introduce \method, a simple and architecture-agnostic method for adapting contrastively pretrained vision-language models, such as CLIP~\cite{radford2021learning}, to few-shot classification. \method{} fine-tunes the vision encoder's projection matrix with Frobenius norm regularization on its deviation from the pretrained weights.
It achieves state-of-the-art performance on 11 few-shot classification benchmarks under both ``few-shot validation''~\citep{huang2024lp++} and ``validation-free''~\citep{silva2024closer} settings. Moreover, by rethinking the non-linear CLIP-Adapter~\cite{gao2024clip} through \method{}’s lens, we design a Regularized Linear Adapter (RLA) that performs better, requires no hyperparameter tuning, is less sensitive to learning rate values, and offers an alternative to \method in black-box scenarios where model weights are inaccessible.
Beyond few-shot classification, \method{} excels in cross-dataset transfer, domain generalization, base-to-new class generalization, and test-time adaptation---where it outperforms prompt tuning while being an order of magnitude faster to train. Code is available at \url{https://github.com/astra-vision/ProLIP}.
\end{abstract}
    
\section{Introduction}
\label{sec:intro}

Contrastive Language-Image Pretraining (CLIP)~\citep{radford2021learning} 
has shown that contrastive learning at scale from noisy image descriptions yields strong visual features. 
CLIP's visual-language space allows for 
a wide range of text-image downstream tasks, including in zero-shot settings. 

At inference, CLIP performs zero-shot classification by measuring the similarity between image embeddings and class labels expressed as text prompts, as shown in~Fig.~\hyperref[fig:teaser_figure]{1a}.
Yet, the performance may still be unsatisfying for data underrepresented in CLIP's pretraining set. Examples of such cases include geospatial data, \textit{e.g.},~EuroSAT~\citep{helber2019eurosat} and specialized data, \textit{e.g.},~FGVCAircraft~\citep{maji2013fine}. This has led to a key challenge in transfer learning: \textit{how to efficiently adapt CLIP to new tasks with only a few labeled samples?}

Using only a few labeled samples, model training risks overfitting.
The common strategy is to avoid full fine-tuning and instead adapt only a few parameters~\citep{kumar2022fine}. Starting from a concept-rich pretrained CLIP model, such parameter-efficient strategies have proved effective for few-shot tasks.
To this end, the literature explores three avenues (\cref{fig:teaser_figure}).
First, linear probing (LP)~\citep{radford2021learning,huang2024lp++,silva2024closer} trains a linear classifier on the frozen visual features. 
Second, Adapters~\citep{gao2024clip,zhang2022tip,song2023meta} introduce a multi-layer perceptron~(MLP) 
on top of frozen visual or text features, using residual connections 
to incorporate zero-shot features. 
Third, prompt tuning~\citep{zhou2022learning,zhou2022conditional} replaces the prompt template with learnable parameters, 
while freezing both vision and text encoders. 

While simple and parameter-efficient, existing CLIP adaptation strategies come with important drawbacks. 
Linear probing~\cite{radford2021learning,lin2023multimodality,huang2024lp++,silva2024closer}, cache-based adapters~\cite{zhang2022tip} and text adapters~\cite{yu2023task} are limited to few-shot settings and cannot handle open-class or cross-dataset tasks. 
Prompt tuning methods~\citep{zhou2022learning,zhou2022conditional,chen2023plot,zhu2023prompt,khattak2023maple} are slow to train 
because gradients must flow through the full text encoder; their performance can also fluctuate 
with context length and class-name positions.
Adapter-based approaches~\citep{song2023meta,gao2024clip,zhang2022tip} 
introduce architectural and hyperparameter sensitivity that complicates robust deployment across datasets.

Compounding these pitfalls, many prior CLIP few-shot adaptation methods also depend on per-dataset validation sets to tune critical hyperparameters~\cite{gondal2024domain,zhang2022tip,gao2024clip,yu2023task,lin2023multimodality}, which conflicts with the spirit of the few-shot setting and undermines the goal of a single, universal configuration. For example, cache-based models~\cite{zhang2022tip,gondal2024domain} typically rely on validation data to choose the cache-logit mixing weight and the scale of intra-modal distance. Recent analyses~\cite{huang2024lp++,silva2024closer} underscore how brittle this practice is: performance drops when hyperparameters are selected under \emph{few-shot validation}~\cite{huang2024lp++} and degrades even more sharply under fully \emph{validation-free} conditions~\cite{silva2024closer}. 

In this paper, we propose a new method for CLIP adaptation, coined \method{}. 
Our approach is simple to implement, requiring only a few lines of code and is highly effective.
It fine-tunes the embedding projection matrix of the vision encoder (which maps visual embeddings into the shared vision-text space) using a cross-entropy loss, while regularizing it to stay close to the pretrained one using the Frobenius norm of their difference. \method{} is a competitive baseline and is robust to variations in training configuration.

\smallskip\noindent\textbf{\method{} offers major advantages:}
\begin{itemize}
    \item It requires no external parameters, eliminating the need for complex architecture searches and hyperparameter tuning. In the \emph{validation-free}~\cite{silva2024closer} setting, it maintains stable performance across learning rates by inversely scaling the regularization loss with the number of support samples per class.
    \item While other methods that fine-tune model weights are limited to ViTs~\cite{khattak2023maple,khattak2023self,zanella2024low}, ProLIP works for both CNNs and ViTs.
    \item Fine-tuning only the projection matrix is highly efficient, taking only $2$ seconds on saved pre-projected features.
    \item It retains CLIP's open-class capability by using native text embeddings as classification weights, unlike~\cite{huang2024lp++,silva2024closer,lin2023multimodality,zhang2022tip,yu2023task} that constrain the classifier to a fixed label set.
     \item Its principle of regularized linear adaptation enables a simplified CLIP-Adapter~\cite{gao2024clip} variant, Regularized Linear Adapter (RLA), which outperforms the original without requiring residual connection and architecture choices.
\end{itemize}

\noindent In experiments, \method performs better or on par when compared to existing methods for few-shot classification, cross-dataset generalization, domain generalization, and base-to-new class generalization. It also significantly outperforms prompt tuning in test-time adaptation, while being one order of magnitude faster to train. Further, \method generalizes seamlessly to other CLIP-like models such as SigLIP~\cite{zhai2023sigmoid}, leading to similar gains.
\section{Related work}
\noindent\textbf{Parameter-efficient fine-tuning (PEFT).} The advent of increasingly larger pretrained vision foundation models with strong generalization enables transfer learning approaches using limited labeled data.
Full fine-tuning is computationally expensive and often underperforms even when compared to linear probing~\citep{kumar2022fine,wei2024stronger}.
PEFT methods adapt models with minimal parameter updates. 
Side-tuning~\citep{zhang2020side} trains a small parallel network to prevent
catastrophic forgetting. 
Optimizing specific parameters, like 
bias terms~\citep{zaken2022bitfit}, is also effective 
but still requires full backpropagation. 
Adapter-tuning methods add adaptation modules to transformer blocks~\citep{houlsby2019parameter, ruckle2021adapterdrop}, but incur higher runtime costs.
LoRA~\citep{hu2022lora} optimizes low-rank matrices in transformer layers to approximate weight changes, significantly reducing the number of parameters to learn.
Prompt-tuning, such as VPT~\citep{jia2022visual}, adds 
learnable prompts to 
input patch embeddings.
However, these methods are 
tailored for transformers and do not directly apply to convolutional networks. 

\medskip\noindent\textbf{Few-shot CLIP adaptation.} 
Inspired by prompt tuning in NLP~\citep{zhong2021factual,li2021prefix}, Zhou \etal~\cite{zhou2022learning} introduced context optimization (CoOp) for vision-language models, but it struggles to generalize to unseen classes. To address this, CoCoOp~\cite{zhou2022conditional} adds a meta-network to generate input-conditional tokens, reducing overfitting to seen classes. Zhu \etal~\cite{zhu2023prompt} noted that unconstrained prompt tuning can overfit in low-shot settings, harming zero-shot performance, and proposed regularizing training by updating only prompts aligned with zero-shot predictions. PLOT~\citep{chen2023plot} uses optimal transport
to match text and visual features, while MaPLe~\cite{khattak2023maple} learns prompts at both input and intermediate layers in vision and text branches with a coupling function.

An alternative to prompt tuning is training a linear probe on top of visual features~\cite{radford2021learning}. Lin \etal~\cite{lin2023multimodality} show that adding text descriptions to the training data improves linear probing, while Huang \etal~\cite{huang2024lp++} blend text embeddings with class-wise learnable classification weights.

Instead of 
prompt learning or 
linear probing, CLIP-Adapter~\citep{gao2024clip} adds an MLP on top of the latent features, with a residual connection to preserve pretrained knowledge. Zhang \etal\cite{zhang2022tip} 
propose a training-free cache-model, converting visual features from the few-shot set into MLP weights. 
Also training-free, Wang \etal~\cite{wang2024a} ensembles zero-shot classifiers with Gaussian Discriminative Analysis (GDA)~\citep{bishop2006pattern}, assuming Gaussian-distributed class features.
CLIPood~\cite{shu2023clipood} fine-tunes the full encoder with beta moving average regularization to maintain similarity to zero-shot weights.
Notably, we only fine-tune the embedding projector, using 0.45\% of the parameters compared to CLIPood and a simpler regularization.
Recently, Zanella \etal~\cite{zanella2024low} showed that applying LoRA~\cite{hu2022lora} to CLIP improves few-shot classification performance, but the method is ViT-specific and causes new class knowledge forgetting~\cite{farina2025rethinking}.

Some works leverage external priors to boost few-shot performance. For instance, APE~\cite{zhu2023not} uses GPT-3 to generate descriptions, CaFo~\cite{zhang2023prompt} uses GPT-3~\cite{brown2020language}, DINO~\cite{caron2021emerging}, and DALL-E~\cite{ramesh2021zero}, while AMU-tuning~\cite{tang2024amu} presents a unified perspective on few-shot adaptation strategies from a logit bias perspective and uses MoCov3~\cite{chen2021empirical} as additional prior. Our work belongs to the category harnessing only the CLIP model with no extra information~\cite{huang2024lp++,silva2024closer,khattak2023maple,zhang2022tip,yu2023task}.

\section{\method}
\label{sec:method}

 CLIP's visual encoder outputs latent visual features, which are then linearly projected into the shared vision-text latent space by a learned linear layer, \ie, visual embedding projector. Thus, the vision encoder $\displaystyle f$ can be written as $\displaystyle f = f_2 \circ f_1$, where $\displaystyle f_2$ represents the linear projection head, and $\displaystyle f_1$ represents all the preceding layers. Given an image $\displaystyle \tI$, this writes:
\begin{equation}
    \label{eq:resnet_processing}
    \vx_o = {f_1}(\tI) , \quad \vv = f_2(\vx_o) = \mW_o \vx_o + \vb_o,
\end{equation}
with $\vx_o \in \mathbb{R}^{D_o}$ the latent visual features before projection, $\mW_o \in \mathbb{R}^{D \times D_o}$ the projection matrix and $\vb_o$ a bias term.
We show that fine-tuning only the projection matrix $\mW_o$ in~\cref{eq:resnet_processing} can be a strong alternative to prompt tuning and feature adapters.
Specifically, the probability that a sample $i$ belongs to the class $k$ is computed as the Softmax over cosine similarities of image-text embeddings:
\begin{equation}
    p_{ik}(\textcolor{black}{\mW_o}) = \frac{\text{exp}\big({(\textcolor{black}{\mW_o} {\vx_o}_i + \vb_o)}^\intercal\vt_k/\tau\big)}{\sum_{j=1}^{K}\text{exp}\big({(\textcolor{black}{\mW_o} {\vx_o}_i + \vb_o)}^\intercal\vt_j/\tau\big)},
\label{eq:proba_ik}
\end{equation}
with $\vt_k$ being the fixed text representation of class $k$ since
the text encoder is frozen, $\tau$ the pretraining temperature parameter, ${\vx_o}_i={f_1}(\tI_i)$ the pre-projection embedding of sample $i$, and $\vb_o$ the frozen bias ($\mathbf{0}$ for ViT backbone). The matrix $\mW_o$ is learned with gradient descent using a cross-entropy loss $L(\mW_o)$: 
\begin{equation}
\label{eq:CE_loss}
    L(\mW_o) = -\frac{1}{NK} \sum_{i=1}^{N} \sum_{k=1}^{K} y_{ik} \log{p_{ik}}(\mW_o),
\end{equation}
where $y_{ik}$ is the ground truth.

\smallskip\noindent\textbf{Regularization.}
Unconstrained fine-tuning can lead to forgetting the rich pretrained knowledge of CLIP. An effective fine-tuning strategy should therefore balance adaptation to the downstream task with preservation of the pretrained representations. Therefore, to prevent significant drift from the original weights (\textit{i.e.}, knowledge forgetting), we add a regularization term based on the Frobenius norm of the difference between the pretrained and fine-tuned matrices. The total loss is:
\begin{equation}
    \text{L}_\text{ProLIP} = L(\mW_o) + \lambda \|\mW_o - \mW_o^{(0)}\|_\text{F}^2,
\label{sec:loss_prolip}
\end{equation}
where $\mW_o^{(0)}$ denotes the pretrained value of $\mW_o$. 

\begin{figure}
    \centering
    \includegraphics[width=1.0\linewidth]{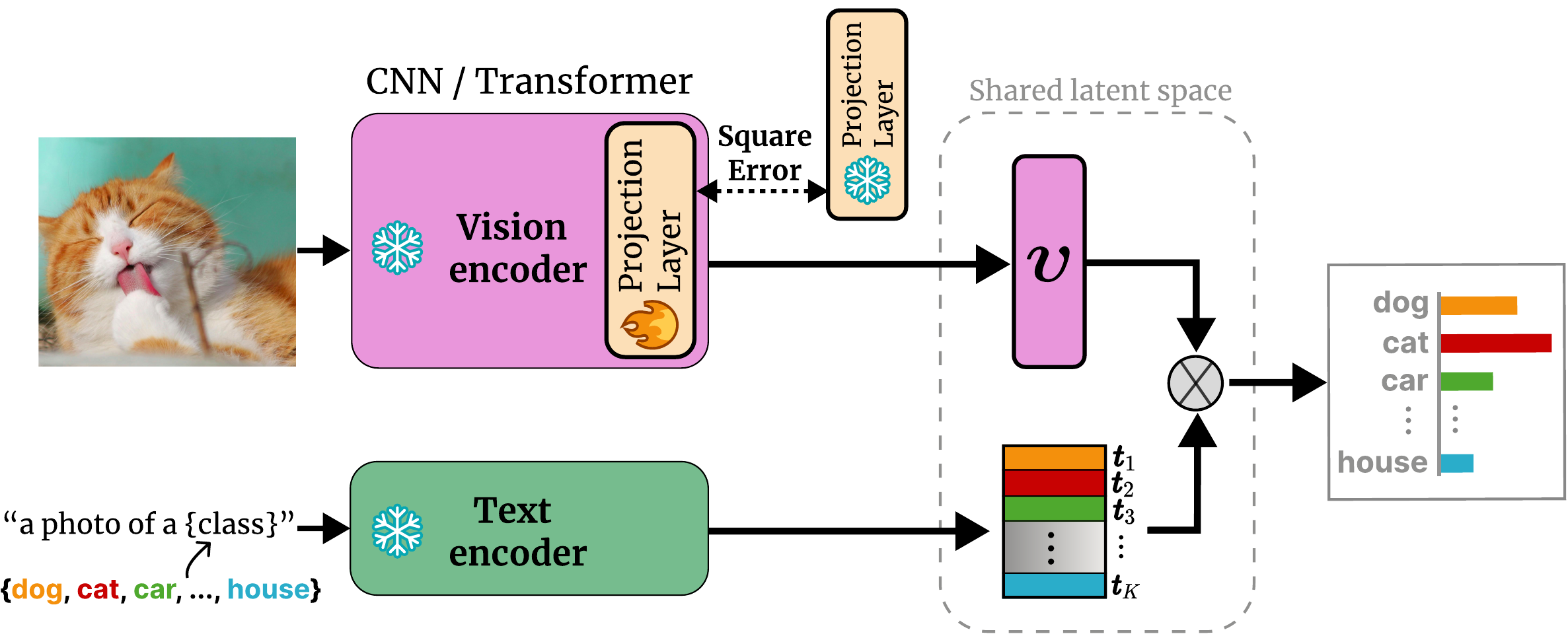}
    \caption{\textbf{\method for few-shot adaptation.} Whether the vision encoder is a CNN or a Transformer, \method trains only the 
    layer that projects the visual embeddings into the shared latent space. The text encoder is frozen, and the text embeddings of the $K$ target concepts are used as classification weights. Training with cross-entropy is regularized by a squared error loss ensuring weights of the projection layer to remain close to pretrained ones.}
    \label{fig:prolip_figure}
\end{figure}

\smallskip\noindent\textbf{$\lambda$ as a function of $N$.} We later show that setting $\lambda$ as a decreasing function of the number $N$ of shots per class (\eg, $\lambda=1/N$ or $\lambda=1/N^2$) helps mitigate overfitting while crucially stabilizing performance across a wide range of learning rates. 
Such observation corroborates classical findings in statistical learning since overfitting risk is known to increase with fewer data points~\citep{hastie2009elements}. Intuitively, in data-scarce settings (\eg, $N=1$), we want to rely more on the zero-shot model for classification by increasing $\lambda$, and when data increases (\eg, $N=16$) we allow the model to learn more from data by decreasing $\lambda$. From learning-theoretic standpoint, this scaling is consistent with generalization bounds where the effective complexity term decreases as N increases~\cite{bartlett2002rademacher}.
An overview of \method{} is shown in \cref{fig:prolip_figure}.

Of note, we are not the first to propose minimizing Frobenius norm as regularization strategy. Such regularization is commonly used to avoid forgetting~\cite{gouk2021distancebased,kirkpatrick2017overcoming}. However, the framework proposed by Gouk \etal~\cite{gouk2021distancebased} is generic and consists of regularized fine-tuning of the model weights across layers, which contradicts PEFT and is shown later in the experimental section to be less effective. \method{} revisits this idea for the projection matrix with a new N-dependent regularization weight $\lambda$ to stabilize few-shot adaptation and preserve zero-shot capability of CLIP. To our knowledge, our strategy has not been done in CLIP adaptation, and offers a stronger and simpler alternative to complex methods.

\method{} can be applied on pre-processed data (\textit{i.e.}, saved pre-projection features), which makes it extremely fast to train. 
A PyTorch-like~\citep{paszke2019pytorch} pseudo-code for \method{} is provided in supplementary material.

\smallskip\noindent\textbf{A logit bias perspective on \method.}
Recently,~\citet{tang2024amu} proposed interpreting CLIP few-shot adaptation methods from a unified perspective of logit bias. That is, every method learns a bias on top of the zero-shot CLIP logits. For instance, 
the bias learned by TaskRes~\cite{yu2023task} is a new linear probe trained on top of frozen visual features $\vv$.
Another example is Tip-Adapter-F~\cite{zhang2022tip} where the learned bias is based on intra-modal similarity measures.
For \method, we fine-tune the projection matrix $\mW_o$. Omitting $\vb_o$ for simplicity, the logits can be written as:
\begin{equation}
    \text{Logits}_{\text{\method}} = \vx_o^\intercal \mW_o^\intercal \vt = \vx_o^\intercal {\mW_o^{(0)}}^{\intercal} \vt + \vx_o^\intercal {\mB}^{\intercal} \vt\,.
\end{equation}
That is, fine-tuning $\mW_o$ is equivalent to learning a matrix $\mB$, initialized with $\mathbf{0}_{D \times D_o}$. Thus, the bias learned by \method is a linear combination of the pre-projected features, trained to match the fixed text-based probe $\vt$.

We show in the experiments that these different biases can be complementary, which suggests that they contain orthogonal knowledge learned during few-shot adaptation.

\section{Revisiting CLIP-Adapter with \method's principles: Regularized Linear Adapter}

\method{} fine-tunes a \emph{linear transformation} of pre-projected features, starting from a zero-shot model as the 
\emph{initialization}, while \emph{regularizing} the weights to stay close to their original values. We revisit here CLIP-Adapter with \method{}'s principles, leading to Regularized Linear Adapter~(RLA) which serves as an alternative to \method in black-box model scenarios.

\smallskip\noindent\textbf{CLIP-Adapter.} CLIP-Adapter learns a randomly initialized MLP $h_\theta(\cdot)$ on top of the frozen vision encoder using a residual connection $\alpha$. The probability that a sample $i$ belongs to the class $k$ in this case writes:
\begin{equation}
    p_{ik}(\textcolor{black}{\theta}) = \frac{\text{exp}\big((\alpha\vv^\intercal + (1-\alpha)[h_\theta(\vv)]^\intercal)\vt_k/\tau\big)}{\sum_{j=1}^{K}\text{exp}\big((\alpha\vv^\intercal + (1-\alpha)[h_\theta(\vv)]^\intercal)\vt_j/\tau\big)},
\label{eq:proba_ik_CLIPADAPTER}
\end{equation}
where $h_{\theta}(\vv) = \texttt{RELU} (\mW_2\texttt{RELU}(\mW_1\vv))$, $\mW_1 \in \mathbb{R}^{D/B \times D}$ and $\mW_2 \in \mathbb{R}^{D \times D/B}$ being the two layers, $B$ the bottleneck dimension and $\texttt{RELU}(x)=\texttt{max}(0,x)$ the activation.

This formulation includes not only the learning rate but also two additional hyperparameters, $\alpha$ and $B$, along with architectural choices such as the number of layers and activation function. This makes the training impractical, as it requires a validation set to select these hyperparameters, which increases computational time. We later corroborate this performance sensitivity in the experimental section through $13\,200$ runs of CLIP-Adapter. 

\begin{figure}
    \centering
    \includegraphics[width=1.0\linewidth]{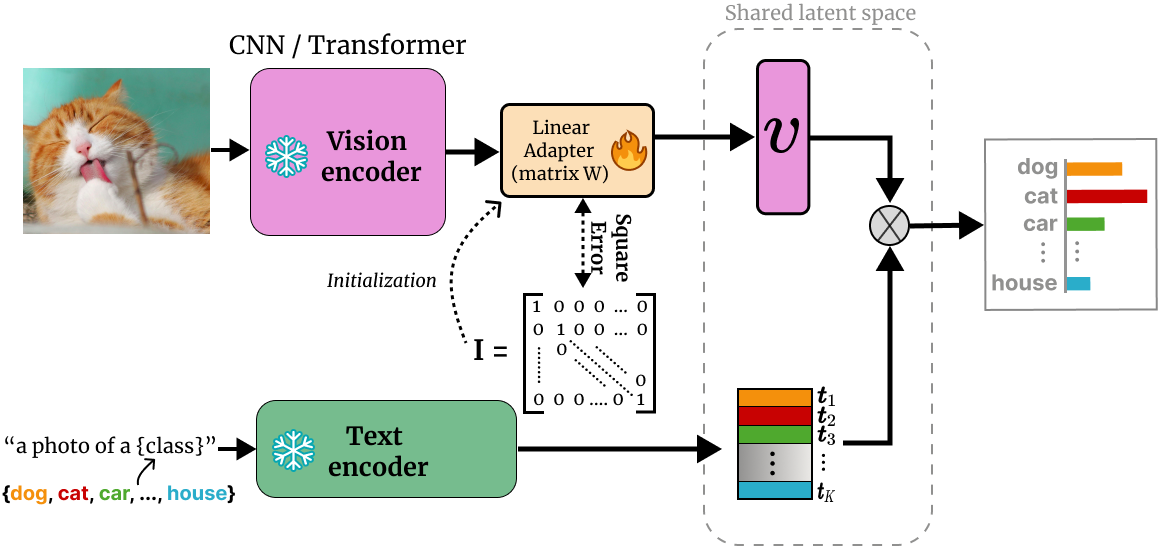}
    \caption{\textbf{Regluarized Linear Adapter (RLA).} RLA is a black-box version of \method{}. Instead of fine-tuning the projection matrix, an external linear adapter is added and trained using cross-entropy loss, with squared-error regularization ensuring that the adapter's weights remain close to the identity matrix.}
    \label{fig:rla_figure}
\end{figure}

\smallskip\noindent\textbf{Regularized Linear Adapter (RLA).} We revisit CLIP-Adapter~\cite{gao2024clip} and incorporate ProLIP's principles by (i) replacing the non-linear MLP with a linear transformation, (ii) initializing it with the identity matrix instead of random weights, and (iii) regularizing it during training with Frobenius norm.
This variant of CLIP-Adapter is called Regularized Linear Adapter (RLA).
For RLA, the probability that a sample $i$ belongs to the class $k$ writes:
\begin{equation}
    p_{ik}(\textcolor{black}{\mW}) = \frac{\text{exp}\big(\vv^\intercal\mW\vt_k/\tau\big)}{\sum_{j=1}^{K}\text{exp}\big(\vv^\intercal\mW\vt_j/\tau\big)},
\label{eq:proba_ik_RLA}
\end{equation}
It can be seen from \cref{eq:proba_ik_RLA} that the linear adapter \mbox{$\mW \in \mathbb{R}^{D\times D}$} is multimodal by construction. 
Following the same principle of \method{}, RLA is trained with cross-entropy loss $L(\mW)$ regularized with the Frobenius norm between the fine-tuned matrix $\mW$ and its initialization $\mI$ (the identity matrix). The total loss of RLA writes: 
\begin{equation}
    \text{L}_\text{RLA} = L(\mW) + \lambda \|\mW - \mI\|_\text{F}^2,
\label{sec:loss_RLA}
\end{equation}
RLA exhibits strong stability across learning rates when $\lambda$ is inversely proportional to $N$, and outperforms the non-linear CLIP-Adapter, challenging the need for a bottleneck layer to learn new features~\cite{gao2024clip}. 
An additional advantage of RLA is that it can substitute for \method{} in black-box settings where the base model's weights cannot be fine-tuned. RLA is illustrated in~\cref{fig:rla_figure}.

\smallskip\noindent\textbf{Conclusive remarks.} \method and RLA suggest CLIP's representations already lie in a space where regularized linear transformations suffice to align features to new tasks. A prominent challenge in selective fine-tuning is choosing the set of parameters to tune for adaptation. 
Our work suggests that these parameters are not necessarily distributed across all the layers, therefore eliminating the need for backpropagation over the entire network~\cite{khattak2023maple,zanella2024low} and paving the way for 
more efficient adaptation of other multimodal foundation models.

\begin{table*}[t]
\begin{center}
\resizebox{0.8\linewidth}{!}{%
\begin{tabular}{l l c c c c c}
\toprule
Method & \# params & $N=1$ & 2 & 4 & 8 & 16 \\
\midrule[0.7pt]
\textcolor{gray}{CLIP (0-shot)} & & \multicolumn{5}{c}{\textcolor{gray}{58.89}}\\
\midrule
\textit{Prompt tuning} \\
\hdashline
CoOp~\cite{zhou2022learning} & $K {\times} M {\times} D_e$ & 59.62\var{3.11} &  63.80\var{2.32} & 67.23\var{1.64} & 71.30\var{0.86} & 74.06\var{0.55} \\
PLOT~\cite{chen2023plot} & $P {\times} K {\times} M {\times} D_e$ & 61.51\var{2.91} & 65.67\var{2.06} & 68.39\var{1.17} & 71.96\var{0.70} & 74.35\var{0.66} \\
KgCoOp~\cite{yao2023visual} & $K {\times} M {\times} D_e$ & 61.36\var{3.04} & 63.23\var{2.06} & 65.73\var{1.15} & 67.50\var{1.11} & 69.01\var{0.79} \\
ProGrad~\cite{zhu2023prompt} & $K {\times} M {\times} D_e$ & 62.46\var{1.89} & 65.88\var{1.46} & 68.52\var{1.15} & 71.82\var{0.11} & 73.95\var{0.68}\\
\midrule
\textit{Adapters} \\
\hdashline
CLIP-Adapter~\cite{gao2024clip} & $2(D_B{\times} D)$ & 60.32\var{0.80} & 61.93\var{0.93} & 65.12\var{0.80} & 69.20\var{0.56} & 72.57\var{0.54}\\
Tip-Adapter-F~\cite{zhang2022tip} & $N {{\times}} K {{\times}} D$ & 61.29\var{0.92} & 62.94\var{0.75} & 66.02\var{0.80} & 69.88\var{0.51} & 73.82\var{0.55}\\
Tip-Adapter-F*~\cite{zhang2022tip} & $N {\times} K {\times} D$ & 63.06\var{1.05} & \second{66.47}\var{0.65} & 68.71\var{0.96} & 71.78\var{1.00} & 74.37\var{0.35} \\
\midrule 
\textit{Linear Probing} \\
\hdashline
LP~\cite{radford2021learning} & $K {\times} D$ & 36.10\var{1.43} & 46.99\var{1.29} & 56.72\var{1.20} & 64.66\var{0.55} & 70.56\var{0.44} \\
LP++~\cite{huang2024lp++} & $K {\times} (D{+}1)$ & \second{63.43}\var{0.90} & 66.20\var{0.72} & \second{69.16}\var{0.79} & \second{72.04}\var{0.46} & \second{74.42}\var{0.45}\\
\midrule
\textit{Model weights} \\
\hdashline
\method & $D_o {\times} D$ & \best{64.21}\var{1.93} & \best{67.43}\var{1.37} & \best{70.58}\var{1.08} & \best{73.73}\var{0.75} & \best{76.50}\var{0.50}\\
\bottomrule
\end{tabular}
}
\caption{\textbf{Few-shot classification \textit{with few-shot validation}.} 
We report the classification accuracy (\%) averaged over 11 datasets and 10 seeds, comparing \method{} to baselines taken from~\cite{huang2024lp++}. 
Note that baselines numbers differ from those reported in the original papers as they used a large validation set to tune hyperparameters. We highlight \best{best} and \second{2nd best}. First row provides zero-shot classification for reference. $D_e$ is the dimension of the token embedding space, $P$ the number of prompts in PLOT, $M$ the context length, and $D_B$ the bottleneck dimension of CLIP-Adapter.}
\label{tab:result_sota}
\end{center}
\end{table*}

\section{Experiments}
\label{sec:exp}

\label{sec:exp_details}
\smallskip\noindent\textbf{Datasets.} Following prior CLIP few-shot learning 
work, we experimentally test \method~on $11$ datasets for few-shot classification and base-to-new generalization: ImageNet~\citep{deng2009imagenet}, SUN397~\citep{xiao2010sun}, DTD~\citep{cimpoi2014describing}, Caltech101~\citep{fei2004learning}, UCF101~\citep{soomro2012ucf101}, Flowers102~\citep{nilsback2008automated}, StanfordCars~\citep{krause20133d}, FGVCAircraft~\citep{maji2013fine}, EuroSAT~\citep{helber2019eurosat}, OxfordPets~\citep{parkhi2012cats} and Food101~\citep{bossard2014food}. For domain generalization experiments we follow ProGrad~\citep{zhu2023prompt}, using ImageNet as source dataset and testing on ImageNet-V2~\citep{recht2019imagenet}, ImageNet-Sketch~\citep{wang2019learning}, ImageNet-A~\citep{hendrycks2021natural} and ImageNet-R~\citep{hendrycks2021many} as out-of-distribution (OOD) datasets. For the cross-dataset transfer experiment, \method is trained on ImageNet and evaluated on the other $10$ datasets, similar to ProGrad~\citep{zhu2023prompt}. For test-time adaptation, we use ImageNet and its 
OOD variants similarly to TPT~\cite{shu2022test}.

\smallskip\noindent\textbf{Training details.}
Following prior work we use \mbox{$N{\in}\{1,2,4,8,16\}$} shots as support training set for few-shot classification. 
For domain generalization and cross-dataset transfer experiments, we use $N{=}4$ like ProGrad~\citep{zhu2023prompt}. For base-to-new generalization, we set $N{=}4$ like ProGrad~\citep{zhu2023prompt} when using ResNet-50 (RN50) and $N{=}16$ like MaPLe~\cite{khattak2023maple} when using ViT-B/16. Unless otherwise stated, we employ ResNet-50 with CLIP weights as the visual encoder, similarly to the literature. 
Training runs for $300$ epochs on a full-batch of features, taking up to 2s on 
one Tesla V100.

\begin{figure*}[t]
\centering

\hfill%
\makebox[0.19\linewidth][c]{\footnotesize$N=1$}%
\makebox[0.19\linewidth][c]{\footnotesize$2$}%
\makebox[0.19\linewidth][c]{\footnotesize$4$}%
\makebox[0.19\linewidth][c]{\footnotesize$8$}%
\makebox[0.19\linewidth][c]{\footnotesize$16$}\\
\includegraphics[height=0.19\linewidth]{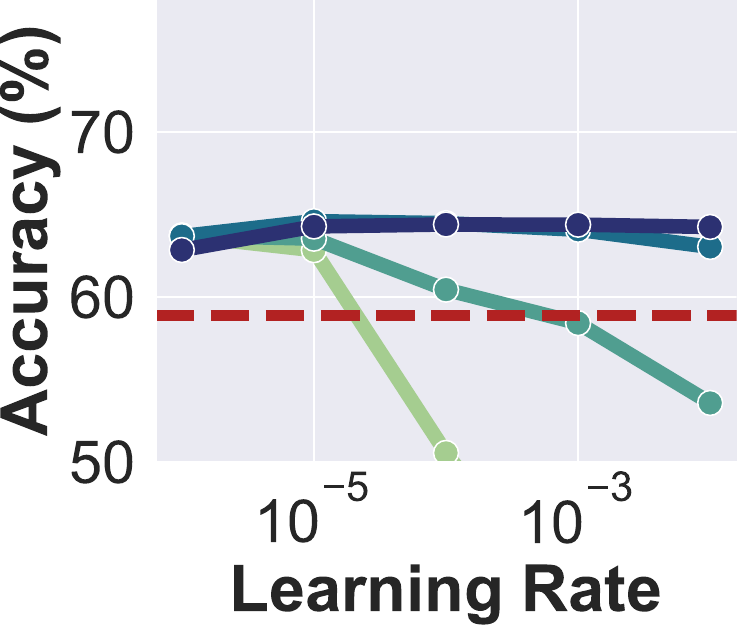}\hfill%
\includegraphics[height=0.19\linewidth]{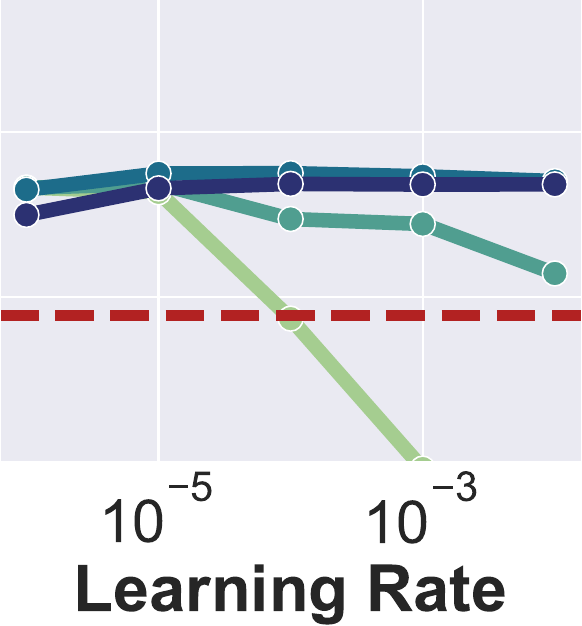}\hfill%
\includegraphics[height=0.19\linewidth]{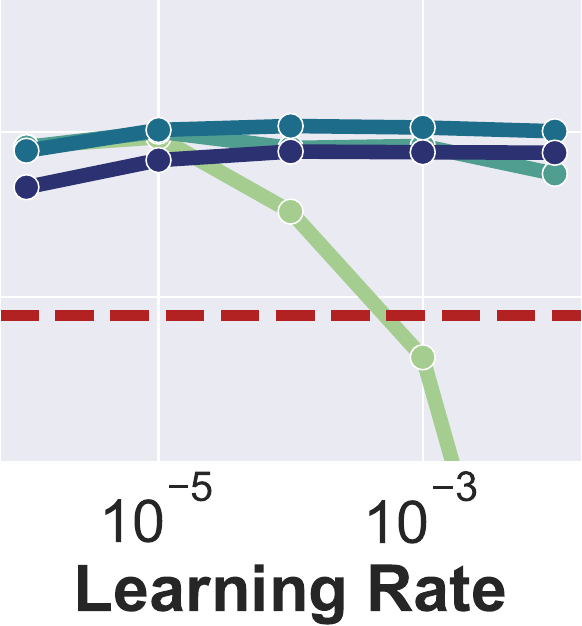}\hfill%
\includegraphics[height=0.19\linewidth]{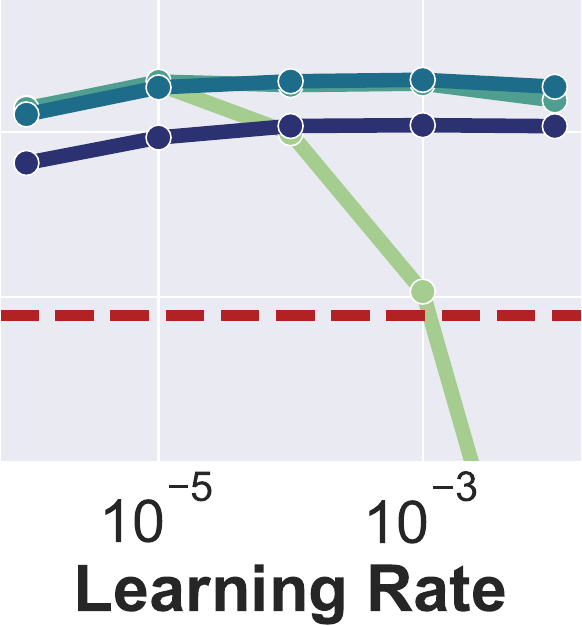}\hfill%
\includegraphics[height=0.19\linewidth]{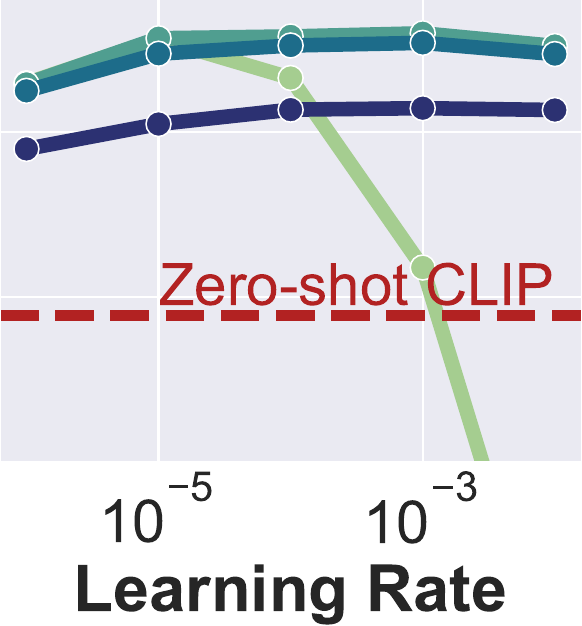}\\
\includegraphics[width=0.7\linewidth]{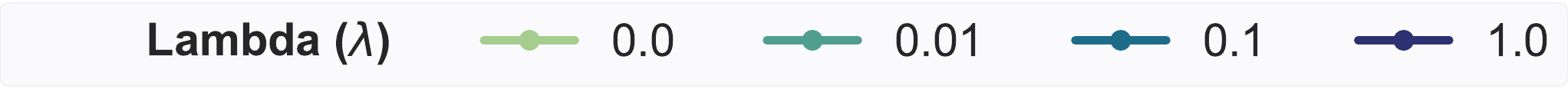}

\caption{\textbf{\method sensitivity to hyperparameter.} Accuracy of \method as function of the hyperparameters (learning rate and regularization weight $\lambda$) for $N\in\{1,2,4,8,16\}$-shot settings. Each data point is an average over 11 datasets and 10 seeds.}
\label{fig:results_reg_vs_no_reg_no_search}
\end{figure*}

\begin{table*}[t]
\newcommand{\varh}[1]{} 

\setlength{\fboxsep}{1pt}
\centering
\begin{subtable}{0.3\linewidth}
    \centering
    \includegraphics[width=1.0\linewidth]{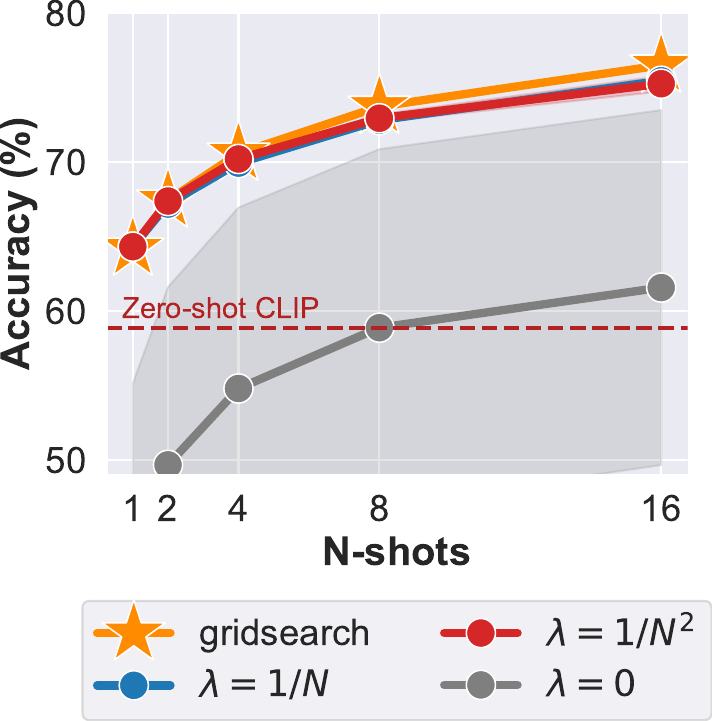}
    \caption{\method variants}
    \label{tab:results_wo_validation_plot}
\end{subtable}%
\hfill%
\begin{subtable}{0.7\linewidth}
\centering
\resizebox{0.7\linewidth}{!}{%
\setlength{\tabcolsep}{0.3em}
\begin{tabular}{l l c c c c c}
\toprule
{Method} & \# params & $N=1$ & 2 & 4 & 8 & 16 \\
\midrule[0.7pt]
\textcolor{gray}{CLIP (0-shot)} & & \multicolumn{5}{c}{\textcolor{gray}{58.89}} \\
\midrule
\textit{Prompt tuning} \\
\hdashline
ProGrad~\cite{zhu2023prompt} & $K {\times} M {\times} D_e$ & 62.61 & 64.90 & 68.45 & 71.41 & 74.28 \\
PLOT~\cite{chen2023plot} & $P {\times} K {\times} M {\times} D_e$ & 62.59 & 65.23 & 68.60 & 71.23 & 73.94 \\
\midrule
\textit{Adapters} \\
\hdashline
TaskRes~\cite{yu2023task} & $K{\times} D$ & 61.44 & 65.26 & 68.35 & 71.66 & 74.42 \\
Tip-Adapter-F~\cite{zhang2022tip} & $N {\times} K {\times} D$ & 60.29 & 62.26 & 65.32 & 68.35 & 71.40 \\
\midrule
\textit{Linear Probing} \\
\hdashline
Cross-modal LP~\cite{lin2023multimodality} & $K {\times} D$ & 62.24 & 64.48 & 66.67 & 70.36 & 73.65 \\ 
CLAP~\cite{silva2024closer} & $K {\times} D$ & \second{62.79} & \second{66.07} & \second{69.13} & \second{72.08} & \second{74.57} \\
\midrule
\textit{Model weights} \\
\hdashline

\methodvalfree & $D_o {\times} D$ & \best{64.33} & \best{67.19} & \best{69.94} & \best{72.82} & \best{75.46} \\
\bottomrule
\end{tabular}%
}%
\caption{Few-shot without validation sets}
\label{tab:results_wo_validation_tab}
\end{subtable}%
\caption{\textbf{Few-shot classification \textit{without validation set}.} \protect\subref{tab:results_wo_validation_plot} \method variants with performance averaged over 11 datasets, 10 seeds, and 4 learning rates $\text{LR} {\in} \{10^{-5},10^{-4},10^{-3},10^{-2}\}$ to study sensitivity. Note the low variance of the parametric formulations ($\lambda=1/N$, $\lambda=1/N^2$) which also reach the performance of the grid search variant --- despite having no access to a validation set. \protect\subref{tab:results_wo_validation_tab} Comparison to validation-free baselines from~\cite{silva2024closer} showing the consistent superiority of \methodvalfree.~\cref{tab:result_sota} defines $D_e, M \& P$.} 
\label{tab:results_wo_validation}
\end{table*}

\smallskip\noindent\textbf{Baselines.} 
We compare against a variety of existing adaptation strategies that harness only the CLIP model without using external pretrained networks.
Among the baselines, we compare to \textit{prompt learning} (\textit{e.g.}, CoOp~\citep{zhou2022learning}, CoCoOp~\cite{zhou2022conditional}, PLOT~\citep{chen2023plot}, KgCoOp~\citep{yao2023visual}, ProGrad~\citep{zhu2023prompt}, MaPLe~\cite{khattak2023maple}), \textit{adapters} (\textit{e.g.}, CLIP-adapter~\citep{gao2024clip}, Tip-adapter~\citep{zhang2022tip}, TaskRes~\cite{yu2023task}), and \textit{linear probing} methods (\textit{e.g.}, LP~\citep{radford2021learning}, LP++~\citep{huang2024lp++}, CLAP~\cite{silva2024closer}). Note that, in \cref{sec:sota_comp}, Tip-adapter performance is reported in two settings following~\citep{huang2024lp++}: Tip-adapter-F where its two crucial hyperparameters are set to $1$ and the validation set is used for early stopping, and Tip-adapter-F* where intensive hyperparameter search is performed to find the best values of the same hyperparameters based on the same validation set. More baselines are included in supplementary material.

\smallskip\noindent\textbf{Fair protocol for hyperparameter tuning.}
Different from the few-shot CLIP literature~\citep{zhang2022tip} relying on large validation sets for hyperparameter tuning, authors of LP++~\citep{huang2024lp++} advocate for a few-shot validation set, \textit{i.e.}, using a validation set with as many shots as in the training set.
Going one step further, we argue that a truly realistic few-shot setting should not use \textit{any} validation set as in~\citep{silva2024closer}.
For comparison purposes, in~\cref{sec:sota_comp} we evaluate in the few-shot validation setting, but the core of our evaluation, from~\cref{sec:realistic} onwards, focuses on the \textit{validation-free setting}.
Moreover, we evaluate \method on 10 random seeds (\textit{i.e.}, support training sets) for each dataset, as advised by~\citet{huang2024lp++}.

For~\cref{sec:sota_comp} only, the learning rate (LR) and regularizer loss weight $\lambda$ are selected by grid search on the few-shot validation set, with \mbox{LR $\in$ $\{10^{-2},10^{-3},10^{-4},10^{-5},10^{-6},10^{-7},10^{-8}\}$} and \mbox{$\lambda \in \{10,1,10^{-1},10^{-2},10^{-3},10^{-4},0\}$}.

For \cref{sec:realistic} and following sections, having no access to a validation set, we show that 
using our regularizer ($\lambda > 0$) prevents severe overfitting, therefore allowing to set a fixed LR over all datasets, and a parametric $\lambda$ as a
decreasing function of the number $N$ of shots.

\subsection{Few-shot classification with few-shot validation}
\label{sec:sota_comp}

\cref{tab:result_sota} reports the average classification accuracy across $11$ datasets and $10$ seeds. Per-dataset performances are in the supplementary material. In all few-shots settings (\textit{i.e.}, \mbox{$N \in \{1,2,4,8,16\}$}), \method clearly outperforms all the baselines, showing a great potential of the extremely simple approach of fine-tuning the visual embedding linear projector with regularization for adaptation.

\smallskip\noindent\textbf{Hyperparameters sensitivity.}
The regularization benefit of \method{} is evident when tested with different hyperparameters fixed across datasets. \cref{fig:results_reg_vs_no_reg_no_search} shows the average accuracy across the same $11$ datasets for $4$ different LRs, combined with regularization ($\lambda \in \{10^{-2},10^{-1},1\}$) or without ($\lambda{=}0$). When $\lambda{=}0$, 
accuracy drops dramatically at higher 
LRs due to overfitting on the few-shot training set, causing drift from the robust pretrained CLIP representation. 
In contrast, using weight regularization ($\lambda>0$) reduces overfitting and LR sensitivity. 
This is supported by grid search statistics (see supp.), 
which show that the best 
LRs span a wide range. Our regularization thus mitigates overfitting, enabling the use of larger LRs 
(\textit{e.g.},~$10^{-2}$). 
This motivates our exploration of a more realistic setting where hyperparameters are not tuned, \textit{i.e.}, \textit{no validation data is used}.

\subsection{Few-shot classification without validation set}
\label{sec:realistic}

An additional merit of~\method{} stems from its lower sensitivity to hyperparameters, as demonstrated in the previous section.
It can be observed from \cref{fig:results_reg_vs_no_reg_no_search} that for lower-shot settings, higher~$\lambda$ values lead to better accuracy, and vice versa.
Therefore, we formulate $\lambda$ as a decreasing function of the number of shots $N$, reporting in \cref{tab:results_wo_validation_plot} the average performance over learning rates $\text{LR} \in \{10^{-5},10^{-4},10^{-3},10^{-2}\}$. It results that our simple parametric formulations of $\lambda$ (\textit{i.e.}, $1/N$, $1/N^2$) lead to almost identical, strong and stable results, competing with our state-of-the-art grid search variant, albeit without the need of a validation set.
A byproduct of our regularizer is the reduced sensitivity to the learning rate (\textit{i.e.}, low variance as seen in~\cref{tab:results_wo_validation_plot}), whereas removing the regularizer (\textit{i.e.}, $\lambda=0$) proves to result in dramatically large variance. 
Detailed numbers for each combination are reported in 
the supplementary material.
Therefore, in~\cref{tab:results_wo_validation_tab} we compare the average across 11 datasets of the validation-free baselines from~\cite{silva2024closer}, and the validation-free \method variant, coined as \methodvalfree, with $\lambda=1/N$ and average over the $4$ tested learning rates. 
The reported performance shows a consistent improvement for any $N$. 

\noindent In terms of parameters, depending on $K$ and the used architecture, \method{} can be more efficient than CLAP and vice versa. For example, \method{} trains 393K/149M (0.0026\%) parameters for ViT-B/16, while CLAP trains 512K/149M (0.0034\%) for ImageNet and 51K/149M (0.00034\%) for Caltech101, resulting in similar percentage range.

\noindent Comparison to architecture-specific baselines (\textit{e.g.}, CLIP-LoRA~\cite{zanella2024low}, normalization techniques~\cite{wang2021tent,zhao2024tuning}) is provided in the supplementary material.

\subsection{Regularized Linear Adapter (RLA)}
\label{sec:RLA}

\begin{figure}
    \centering
    \includegraphics[width=1.0\linewidth]{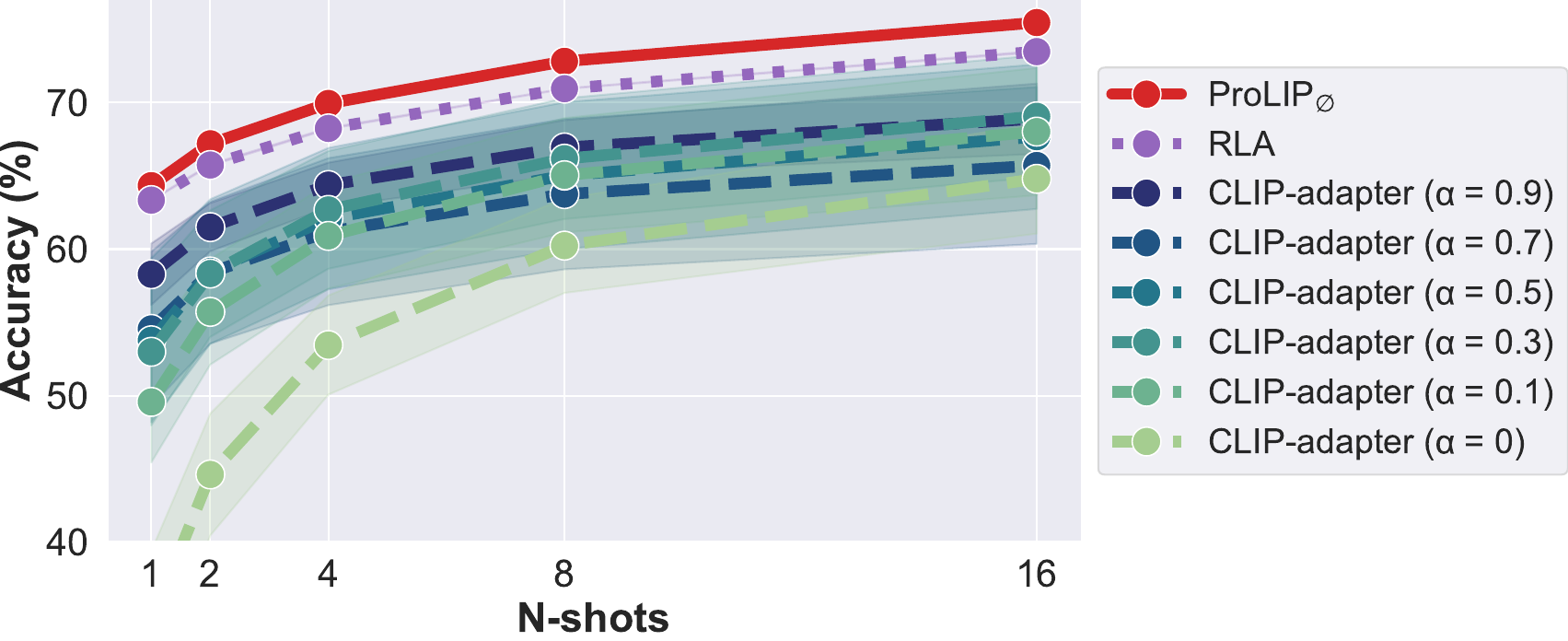}
    \caption{\textbf{Improving CLIP-Adapter with ProLIP's principles} results in Regularized Linear Adapter (RLA) variant. We report classification accuracy (\%) averaged over 11 datasets, 10 seeds, and 4 learning rates $\text{LR} {\in} \{10^{-5},10^{-4},10^{-3},10^{-2}\}$ for CLIP-Adapter with different $\alpha$ values, \methodvalfree and RLA with $\lambda=1/N$.
    Variance is halved for readability.
    }
    \label{fig:linear_adapter}
\end{figure}

\cref{fig:linear_adapter} illustrates the sensitivity of CLIP-Adapter to learning rate (LR) and residual weight ($\alpha$) through \textbf{13\,200 training runs}~(11 datasets $\times$ 10 seeds $\times$ 5 settings $\times$ 4 LRs $\times$ 6 $\alpha$). Our RLA outperforms CLIP-adapter, showing stability across different LRs in the \emph{validation-free} setting, though it still trails behind \methodvalfree. Detailed LR results are provided in the supplementary material. These findings reinforce the core principles of \method{} and shed further light on why it is so effective, apart from the perhaps surprising effect of fine-tuning the visual embedding projector.

\subsection{Generalization of few-shot models}

Achieving generalization in a few-shot framework is challenging but crucial for evaluating the practical use of few-shot methods. We here explore three aspects of generalization: domain generalization, cross-dataset generalization and base-to-new generalization.
Comparison is done only among the few-shot methods; the zero-shot CLIP performance is included as reference.
For \methodvalfree, we systematically use $\lambda=1/N$ and a fixed LR of $10^{-5}$.

\smallskip\noindent\textbf{Cross-dataset generalization.}
\cref{tab:cross_dataset} shows \methodvalfree outperforms ProGrad on $6$ out of $11$ datasets, and on average, but zero-shot CLIP remains the strongest baseline. As argued in CoCoOp~\cite{zhou2022conditional}, training on ImageNet leads to good generalization to datasets like OxfordPets and Caltech101. However, for fine-grained or specialized datasets (\textit{e.g.}, FGVCAircraft, DTD), \methodvalfree outperforms other methods but still lags behind zero-shot performance. Overall, \methodvalfree retains the most zero-shot capability and shows better cross-dataset transferability.

\begin{table}[t]
\centering
\setlength{\tabcolsep}{0.3em}
\newcommand{\db}[1]{\rot{#1}}
\resizebox{1.0\linewidth}{!}{%
\begin{tabular}{lcccccccccccc}
\toprule
& Source & \multicolumn{9}{c}{Target}&\\
\cmidrule{2-2}\cmidrule(l){3-12}
Method & \db{ImageNet} & \db{Caltech101} & \db{OxfordPets} &\db{StanfordCars} & \db{Flowers102} & \db{Food101} & \db{FGVCAircraft} & \db{SUN397} & \db{DTD} & \db{Eurosat} & \db{UCF101} & Average \\
\midrule
\textcolor{gray}{CLIP (0-shot)} & \textcolor{gray}{60.35} & \textcolor{gray}{85.84} & \textcolor{gray}{85.75} & \textcolor{gray}{55.78} & \textcolor{gray}{65.98} & \textcolor{gray}{77.35} & \textcolor{gray}{17.07} & \textcolor{gray}{58.85} & \textcolor{gray}{42.69} & \textcolor{gray}{36.22} & \textcolor{gray}{61.80} & \textcolor{gray}{58.88}\\
\midrule
CoOp & 61.34 & 84.48 & 85.99 & 54.16 & 60.10 & \second{75.48} & 14.09 & 57.48 & 35.32 & \second{26.72} & 57.56 & 55.70\\
CoCoOp & 61.04 & 84.73 & \underline{86.42} & 52.34 & 61.24 & 73.79 & 13.74 & 55.94 & 36.60 & 23.46 & 57.97 & 55.21 \\
Prograd & \second{62.17} & \best{88.30} & \best{86.43} & \best{55.61} & \second{62.69} & \best{76.76} & \second{15.76} & \best{60.16} & \second{39.48} & 24.87 & \second{58.70} & \second{57.36}\\

{\methodvalfree} & \best{62.55} & \second{86.99} & {84.00} & \second{54.19} & \best{64.03} & {74.95} & \best{16.99} & \second{59.67} & \best{41.09} & \best{36.55} & \best{60.95} & \best{58.36}\\

\bottomrule
\end{tabular}%
}
\caption{\textbf{Cross-dataset generalization.} Training is performed on 4-shot ImageNet (source), except for `CLIP'  which is 0-shot. The learned models are evaluated on $10$ other datasets (target). Baselines' scores are average of 3 seeds reported from ProGrad~\citep{zhu2023prompt}. 
}
\label{tab:cross_dataset}
\end{table}

\smallskip\noindent\textbf{Domain generalization (DG).}
In this setting, the set of classes is fixed in both in-domain and OOD datasets.
Following ProGrad, we train \methodvalfree on ImageNet (IN) as source dataset (with 
$N{=}4$), and assess it on ImageNet-V2 (IN-V2), ImageNet-Sketch (IN-S), ImageNet-A (IN-A) and ImageNet-R (IN-R). \cref{tab:result_DG} shows that \methodvalfree is on par with or better than other methods on source and especially OOD domains, for both ResNet and ViT CLIP backbones.
Of note, \methodvalfree's performance stability across learning rates also applies for DG. For instance, with RN50, increasing LR to $10^{-2}$ results in an OOD accuracy drop of only -0.18\% for \methodvalfree (corroborating stability results in few-shot classification, cf.~\cref{tab:results_wo_validation_plot}), while Tip-adapter-F and TaskRes dramatically drop by \textbf{-2.09\%} and \textbf{-10.21\%}, resp.

\begin{table}
    \centering
        \resizebox{0.90\linewidth}{!}{%
        \newcolumntype{H}{@{}>{\setbox0=\hbox\bgroup}c<{\egroup}@{}}
        \setlength{\tabcolsep}{0.6em}
        \begin{tabular}{l c c c c c c c c}
        \toprule
        Method & \multicolumn{2}{c}{RN50} & \multicolumn{2}{c}{RN101} & \multicolumn{2}{c}{ViT-B/16} & \multicolumn{2}{c}{ViT-B/32}\\
        \cmidrule(lr){2-3}\cmidrule(lr){4-5}\cmidrule(lr){6-7}\cmidrule(lr){8-9}
        & IN & OOD & IN & OOD & IN & OOD & IN & OOD \\
        \midrule[0.7pt]
        \textcolor{gray}{CLIP (0-shot)}  &  \textcolor{gray}{60.34}  &  \textcolor{gray}{43.31}  &  \textcolor{gray}{61.24}  &  \textcolor{gray}{48.71}  &  \textcolor{gray}{68.79} &  \textcolor{gray}{59.87}  & \textcolor{gray}{62.00} &  \textcolor{gray}{50.06}\\  
        \midrule
        LP  &  41.29  &  21.19  &  47.01  &  28.33  &  54.70 &  35.09  &  46.77  &  28.81\\ 

        CoOp  &  61.34  &  40.84  & 63.99 &  47.48  & 69.86 &  58.32  &  64.74 &  48.06\\

        CoCoOp  &  61.04  &  40.42  & 63.59  &  47.34  & 70.13  &  58.17  & 64.63  &  47.93\\

        Prograd  & {62.17} &  {42.23}  &  {64.98}  &  {48.53}  & {70.45}  &  {59.05}  &  {65.36}  &  {49.39} \\
        
        TaskRes 
        & \best{62.61} &  \second{42.55}  &  \best{65.57}  & 48.19  & \best{71.01} & \second{59.59}  &  \second{65.99}  &  49.63 \\
        Tip-Adapter-F 
        & 60.88 &  41.72  &  64.85  & \second{48.64}  & 70.17  &  59.04  &  65.63  &  \best{49.97} \\

        {\methodvalfree} & \second{62.55} & \best{43.20} & \second{65.11} & \best{48.81} & \second{70.94} & \best{59.97} & \best{66.00} & \second{49.86} \\
        \bottomrule
        
        \end{tabular}
        }
        \caption{\textbf{Domain generalization.} 4-shot training on ImageNet~(source) and evaluation on OOD variants (IN-V2, IV-S, IN-A, IN-R) with different visual backbones. We report accuracy over source ImageNet (IN) along with the average only over OOD variants to show generalization (`OOD'). All baselines 
        are reported from ProGrad~\citep{zhu2023prompt} except for TaskRes and Tip-Adapter-F which we re-implemented.
    } 
    \label{tab:result_DG}
        
\end{table}

\smallskip\noindent\textbf{Base-to-new generalization.} In this setting, we divide all classes into two groups: base and new classes. Training is performed on base classes and testing on both base and new ones.
The harmonic mean is reported to assess the trade-off.
In \cref{tab:base_to_new_res}, we see that \methodvalfree significantly outperforms ProGrad~\cite{zhu2023prompt} in \textit{total harmonic mean} across 11 datasets. Additionally, \methodvalfree is competitive with MaPLe~\cite{khattak2023maple}, a method specifically designed for few-shot generalization. For comparison fairness, we report the three variants of MaPLe: MaPLe{$\ddagger$} trains $9 {\times}$ more parameters than \method ($3.55$M vs. $0.39$M);  MaPLe{$\dagger$} trains a similar number of parameters ($0.41$M); 
MaPLe{*} is a shallow version that trains prompts only on the first layer of vision and language branches. Our method is architecture agnostic, while all versions of MaPLe work only on ViTs and require backpropagation over the entire vision and text encoders.

\begin{table}[t]
    \centering
    \vspace{-2mm}
    
    \begin{subtable}[t]{0.495\linewidth}
        \centering
        
        \resizebox{0.95\textwidth}{!}{%
        \begin{tabular}{l cc|c}
            \toprule
            & Base & New & $\text{H}$ \\
            \midrule
            \textcolor{gray}{CLIP} & \textcolor{gray}{61.72} & \textcolor{gray}{65.91} & \textcolor{gray}{63.75} \\
            \midrule
            CoOp & 71.96 & 61.26 & 66.18 \\
            CoCoOp & 72.23 & 60.77 & 66.01 \\
            ProGrad & \second{73.29} & \second{65.96} & \second{69.43} \\
            {\methodvalfree} & \best{75.45} & \best{69.43} & \best{72.31} \\
            \bottomrule
        \end{tabular}
        }
        \caption{ResNet-50}
    \end{subtable}%
    \hfill
    \begin{subtable}[t]{0.495\linewidth}
        \centering
        \resizebox{0.95\textwidth}{!}{%
        \begin{tabular}{l cc|c}
            \toprule
            & Base & New & $\text{H}$ \\
            \midrule
            \textcolor{gray}{CLIP} & \textcolor{gray}{69.34} & \textcolor{gray}{74.22} & \textcolor{gray}{71.70} \\
            \midrule
            CoOp & \second{82.69} & 63.22 & 71.66 \\
            CoCoOp & 80.47 & 71.69 & 75.83 \\
            {MaPLe{*}} & {80.10} & {73.52} & {76.67} \\
            {MaPLe{$\dagger$}} & {82.29} & {74.34} & {78.11} \\
            {MaPLe{$\ddagger$}} & 82.28 & \best{75.14} &  \second{78.55} \\
            \methodvalfree & \best{83.85} &	\second{74.78} & \best{79.08} \\
            \bottomrule
        \end{tabular}
        }
        \caption{ViT-B/16}
    \end{subtable}
\caption{\textbf{Base-to-new.} Performance comparison of methods on ResNet-50 and ViT-B/16 architectures across 11 datasets.
}
\label{tab:base_to_new_res}
\end{table}

\subsection{Analysis and Discussion}
\label{sec:ablations}

\smallskip\noindent\textbf{Comparison to full and last layer fine-tuning.}
We compare \methodvalfree with full fine-tuning of the visual backbone. Results in \cref{tab:results_full_tuning} show that full fine-tuning is 
far behind \methodvalfree, and even degrades zero-shot performance for $N=1,2$ and 4-shots. 
LR is $10^{-5}$ for these experiments, and \methodvalfree is shown for different $\lambda$ values (including $\lambda=0$). These results confirm that full fine-tuning faces a high risk of overfitting especially in low-shot regimes, advocating for PEFT 
methods like \method.

Moreover, we show the results of fine-tuning the last layer (\textit{i.e.}, the attention pooling layer) of the backbone. For the same 
LR${=}10^{-5}$, the performance lags behind \methodvalfree, with $8\times$ more trainable parameters. Importantly, we also add results of last-layer fine-tuning when we increase 
LR to $10^{-4}$, showing dramatically decreased performance, especially for extremely low-shot setting (\textit{e.g.}, 1-shot).
\begin{table}[t]
\centering
\resizebox{0.95\linewidth}{!}{%
\begin{tabular}{l ccc c c c c}
\toprule
\multicolumn{1}{c}{Method} & \# params & $N=1$ & 2 & 4 & 8 & 16 \\
\midrule[0.7pt]
\textcolor{gray}{CLIP (0-shot)} & - & \multicolumn{5}{c}{\textcolor{gray}{58.89}} \\
\midrule
Full Fine-tuning & 38.32M &  46.09 & 51.85 & 58.06 & 62.22 & 67.74 \\
\midrule
Last layer FT ($10^{-5}$) & 14.79M &  61.26 &  64.37 & 67.99 & 71.58 & 75.53  \\
Last layer FT ($10^{-4}$) & 14.79M & 47.47 & 54.84 & 62.69 & 68.98 & 74.56 \\
\midrule
\methodvalfree ($\lambda= 0$) & 2.10M &  \second{62.84} & 66.35 & \second{69.69} & \second{72.89} & \second{75.65} \\
\methodvalfree ($\lambda= 1/N$) & 2.10M &  \best{64.28} & \second{67.07} & 69.68 & 72.57 & 75.20 \\ 
\methodvalfree ($\lambda= 1/N^2$) & 2.10M &  \best{64.28} & \best{67.32} & \best{70.22} & \best{73.10} & \best{75.68} \\
\bottomrule

\end{tabular}%
}
\caption{\textbf{Comparison to full fine-tuning.} We report the classification accuracy (\%) averaged over 11 datasets, comparing \methodvalfree to full fine-tuning of the vision encoder, and fine-tuning (`FT') only the last layer, \textit{i.e.} the attention pooling layer.}
\label{tab:results_full_tuning}
\end{table}

\noindent{\textbf{Complementarity to other methods.}} 
~\cref{tab:compl} corroborates 
the complementarity of \method with other methods, showing the benefit of combining \methodvalfree with either TaskRes or Tip-Adapter-F.
We argue, from the same perspective of logit bias discussed in~\cite{tang2024amu}, that each of these methods learns a specific bias on top of zero-shot CLIP, and that these biases contain orthogonal information. For instance, TaskRes learns an element-wise adapter on top of the text embeddings (\textit{i.e.}, the classifier weights), while Tip-Adapter-F learns an adapter initialized with intra-modal similarities (\textit{i.e.}, cache model).
\method's learned bias stems from re-leveraging the pre-projection features to create new combinations adapted to the fixed probe. Interestingly, only a marginal improvement is observed when combining \method and RLA, which suggests that they are best viewed as alternatives---an expected outcome since both methods learn linear transformations of visual features, leading to redundancies when combined together.
More details are provided in the supplementary material.

\noindent 
\textbf{\method with other CLIP-like models.} To verify the generalizability of \method to other CLIP-like models, we provide in~\cref{tab:siglip_results} average results across 11 datasets for \method with SigLIP~\cite{zhai2023sigmoid} ViT-B/16. Despite high zero-shot accuracy (+16.37\% higher than zero-shot CLIP using RN50), \method further advances the accuracy for all few-shot settings. In contrast with CLIP, which benefits from augmented views of the data, we surprisingly found that for SigLIP, adding more views does not improve the accuracy so we employed only one view. Subsequently, saving pre-projected features is extremely fast (a maximum of $57$ sec. on a single V100-32GB GPU for 16-shot ImageNet), with training taking only around $7$ sec. Moreover, in the validation-free setting, \methodvalfree exhibits remarkable robustness to the learning rate (see standard deviation values in~\cref{tab:siglip_results}), which aligns with our previous results with CLIP (\cref{tab:results_wo_validation_plot}).

\noindent We believe \method can be further applied to other domains, like PointCLIPV2~\cite{zhu2023pointclip} for 3D point cloud few-shot classification. \method might relax the need for weight combination of the logits of multiple views, as this knowledge can be internalized in the projection weights by showing different views across iterations.

\begin{table}[t]
\centering
\resizebox{1.\linewidth}{!}{%
\begin{tabular}{l ccc c c c c}
\toprule
\multicolumn{1}{c}{Method}  & $N=1$ & 2 & 4 & 8 & 16 \\
\midrule[0.7pt]
\textcolor{gray}{SigLIP (0-shot)} & \multicolumn{5}{c}{\textcolor{gray}{75.26}} \\
\midrule
\method (grid search)& \best{78.28} & \best{80.49} & \best{82.29} & \best{84.10} & \best{85.82} \\

\midrule
\methodvalfree ($\lambda= 1/N$) & \second{77.45}\var{0.00} & 78.70\var{0.00} & 80.06\var{0.01} & 81.79\var{0.04} & 83.55\var{0.06} \\
\methodvalfree ($\lambda= 1/N^2$) & \second{77.45}\var{0.00} & \second{79.38}\var{0.00} & \second{81.50}\var{0.01} & \second{83.68}\var{0.03} & \second{85.35}\var{0.03}\\
\bottomrule

\end{tabular}%
}
\caption{\textbf{\method with SigLIP.} We report the classification accuracy (\%) averaged over 11 datasets and 10 seeds for \method and \methodvalfree with SigLIP ViT-B/16~\cite{zhai2023sigmoid}. For \methodvalfree, we report average and standard deviation over 3 $\text{LR} {\in} \{10^{-4},10^{-3},10^{-2}\}$.} 
\label{tab:siglip_results}
\end{table}

\begin{table}[t]
\newcommand{\varh}[1]{} 

\setlength{\fboxsep}{1pt}
\centering
\resizebox{0.9\linewidth}{!}{%
\setlength{\tabcolsep}{0.3em}
\begin{tabular}{l c c c c c}
\toprule
{Method} & $N=1$ & 2 & 4 & 8 & 16 \\
\midrule[0.7pt]
\textcolor{gray}{CLIP (0-shot)} & \multicolumn{5}{c}{\textcolor{gray}{58.89}} \\
\midrule
\methodvalfree & 64.40 & 67.28 & 70.08 & 72.97 & 75.57 \\
\methodvalfree + RLA & 64.47 & 67.38 & 70.19 & 73.02 & 75.65 \\
\methodvalfree + Tip-Adapter-F~\cite{zhang2022tip} & 64.53 & 67.47 & 70.30 & 73.23 & 75.89 \\ 
\methodvalfree + TaskRes~\cite{yu2023task} & \best{65.01} & \best{68.18} & \best{71.00} & \best{73.78} & \best{76.23} \\
\bottomrule
\end{tabular}%
}%
\caption{\textbf{Complementarity to other methods.} We report the classification accuracy (\%) averaged over 11 datasets and 10 seeds. LR is fixed to $10^{-4}$ for all datasets, and $\lambda=1/N$.}
\label{tab:compl}
\end{table}

\subsection{\method for Test-time Adaptation}
In this section, our goal is to show that \method can be applied beyond supervised few-shot CLIP adaptation. Motivated by the risk of ``overfitting'' the source domain in classic prompt tuning methods~\citep{zhou2022learning,zhou2022conditional}, Shu \etal~\cite{shu2022test} pioneered test-time prompt tuning (TPT), aiming to learn adaptive prompts on the fly using a single test image. 

\smallskip\noindent{\textbf{TPT background knowledge.}} TPT aims to learn a context specific to each test image in an unsupervised way. Given an unlabeled test image $\tI_\text{test}$, 
the prompt is learned by minimizing the average prediction entropy over different augmented views of $\tI_\text{test}$. Moreover, \textit{confidence selection} filters out the augmented views with high entropy predictions, which might lack important information for classification. More details are provided in the supplementary material.

\begin{table}[t]
\centering
\resizebox{1.\linewidth}{!}{%
\begin{tabular}{l c c c c c c c}
\toprule
Method & IN & IN-A & IN-V2 & IN-R & IN-S & Average & Avg. OOD \\
\midrule
\textcolor{gray}{CLIP (0-shot)} & \textcolor{gray}{60.33} & \textcolor{gray}{23.79} & \textcolor{gray}{53.31} & \textcolor{gray}{60.58} & \textcolor{gray}{35.46} & \textcolor{gray}{46.69} & \textcolor{gray}{43.29}\\
\midrule
\multicolumn{2}{l}{\textit{w/o few-shot training on} IN} \\
\hdashline
TPT~\cite{shu2022test}  & \second{60.74} & \second{26.67} & \second{54.70} & \second{59.11} & \second{35.09} & \second{47.26} & \second{43.89}  \\
\method$_{\!\!\text{test-time}}$ & \best{62.00} & \best{33.76} & \best{56.03} & \best{62.69} & \best{37.29} & \best{50.35} & \best{47.44} \\
\midrule
\multicolumn{2}{l}{\textit{w/ 16-shot training on} IN} \\
\hdashline
CoOp~\cite{shu2022test} & {63.33} & 23.06 & 55.40 & 56.60 & 34.67 & 46.61 & 42.43 \\
TPT + CoOp~\cite{shu2022test} & \second{64.73} & \second{30.32} & \second{57.83} & {58.99} & \second{35.86} & \second{49.55} & \second{45.75} \\
\method & 64.48 & 22.75 & 56.24 & \second{59.56} & 34.80 & 47.57 & 43.34 \\
\method$_{\!\!\text{test-time}}$ + \method & \best{66.90} & \best{32.96} & \best{58.77} & \best{61.78} & \best{36.97} & \best{51.48} & \best{47.62} \\
\bottomrule
\end{tabular}}
\caption{\textbf{Robustness to natural distribution shifts in test-time adaptation.}
Experiments are done with RN50 backbone, without few-shot training on IN (top) and with 16-shot training (bottom).  
}
\label{tab:test_time}
\end{table}

\smallskip\noindent{\textbf{Test-time \method.}} We do not introduce a new way for CLIP test-time adaptation but simply follow the same experimental setting as TPT (\textit{i.e.}, 1-step entropy minimization of averaged prediction probability distribution, confidence selection), although \method optimizes the projection weight matrix $\mW_o$ instead of the prompt as in TPT.
We name this~\method variant as~\method$_{\!\!\text{test-time}}$.
\cref{tab:test_time} shows that \mbox{\method$_{\!\!\text{test-time}}$} yields superior results to TPT on ImageNet and natural distribution shifts, while being 
one order of magnitude faster to train.
For direct comparison, we separate methods that perform 16-shot training on ImageNet.
Of note, even without few-shot training, \method$_{\!\!\text{test-time}}$ still outperforms CoOp and TPT+CoOp.
We further advance \method$_{\!\!\text{test-time}}$ results with 16-shot training.
\section{Conclusion}
We propose a simple and efficient method for adapting CLIP to few-shot classification by fine-tuning the visual projection matrix.
Moreover, we show advantages of including a Frobenius norm regularizer: it prevents the drift from pretrained weights 
and improves robustness to hyperparameter choice, thus making our method an appealing approach to practical few-shot adaptation. Additionally, we reflect on the practice of using non-linear adapters from our method's perspective and propose a strong and robust regularized linear adapter. We provide evidence of the competitiveness of \method in few-shot classification, 
generalization and test-time adaptation, as well as its complementarity with other methods,
rendering it a potential general framework for further applications.

\smallskip\noindent{\textbf{Acknowledgment.} This work was partially funded by French project SIGHT (ANR-20-CE23-0016). It was performed using HPC resources from GENCI–IDRIS (Grants
AD011014477R1, AD011012808R3). The authors thank Clément Weinreich for insightful discussion.}
{
    \small
    \bibliographystyle{ieeenat_fullname}
    \bibliography{main}
}
\clearpage
\maketitlesupplementary

\normalsize
This document provides:
\begin{itemize}
    \item A PyTorch-like pseudo-code for \method{}, shown in~\cref{alg:prolip_pytorch}.
    \item {Per-dataset performance of few-shot classification {with few-shot validation} in~\cref{sec:per-dataset}, complementing~\cref{tab:result_sota}.}
    \item {Grid search and hyperparameter sensitivity in~\cref{sec:grid_search_hyp}, as well as the data of~\cref{tab:results_wo_validation_plot},~\cref{tab:results_wo_validation_tab} and \cref{fig:results_reg_vs_no_reg_no_search}.}
    \item {Base-to-new generalization detailed performance in \cref{sec:app_base_to_new}.}
    \item {Experiments on fine-tuning the text embedding projection matrix in~\cref{sec:text_embed_proj}.}
    \item {Test-time adaptation details in \cref{sec:app_testtime}.}
    \item {Additional comparison of \methodvalfree to architecture-specific methods in~\cref{sec:arch_spec}.}
    \item {Details about RLA, complementarity analysis, and additional experiments and ablations in~\cref{sec:further_analysis}.}
    \item {Training of \method in~\cref{sec:add_training_details}.}
    \item {Preliminaries on CLIP in~\cref{sec:preliminaries_supp}.}
\end{itemize}

\begin{algorithm}[t]
\caption{PyTorch-like pseudo-code for \method.}
\label{alg:prolip_pytorch}

\lstset{
    language=Python,
    basicstyle=\bfseries\color[rgb]{0,0,0}\ttfamily\scriptsize,  
    commentstyle=\color[rgb]{0,0.5,0},
}

\begin{lstlisting}[language=python]
# target: Ground truth
# lmda: regularization loss weight
# Wo : Pretrained projection matrix
# bo : Pretrained bias term (only ResNet, 0 for ViT)
# xo:  output visual embeddings (N*K, Do)
# text_weights: normalized embeddings of classnames (K,D)

# Copy initial weights for use in the regularization loss
Wo_0 = copy.deepcopy(Wo)
# Set embedding projection matrix as trainable weights
Wo.requires_grad = True
bo.requires_grad = False

v = xo @ Wo + bo
v = l2_normalize(v,dim=-1) 

#compute the cosine similarity scores
logits = 100. * v @ text_weights.T

#compute regularized loss
SE_loss = nn.MSELoss(reduction='sum')
loss = CE_loss(logits, target) + lmda * SE_loss(Wo, Wo_0)

\end{lstlisting}
\end{algorithm}

\section{Details on few-shot classification with few-shot validation}
\label{sec:per-dataset}
In addition to the average across datasets in \cref{tab:result_sota}, Tabs.~\ref{tab:per-dataset-perf-1}-\ref{tab:tab:per-dataset-perf-2} provide the \textit{per-dataset performance} of all methods, with for each the average accuracy over $10$ seeds (\textit{i.e.}, support sets). \method performs particularily well on DTD, UCF101, StanfordCars, FGVCAircraft and EuroSAT. For some specific settings, \textit{e.g.}, 1-shot DTD, 
16-shot StanfordCars, 8 and 16-shot FGVCAircraft, the improvements over state-of-the-art are significant. On the other hand, for datasets like OxfordPets and Food101, where the zero-shot performance is already good, \method and other baselines are outperformed by prompt learning methods (\textit{e.g.}, ProGrad). This might be due to the relatively lower number of parameters in the latter, making them less prone to overfitting in very low-shot settings; when the number of shots increases, \textit{e.g.}, 8-16 shots, \method and prompt learning perform on par.

Future research may include the zero-shot accuracy on the few-shot training set in the parametric formulation of the regularization loss weight (\textit{i.e.}, $\lambda$). That is, the higher the zero-shot accuracy, the 
smaller should be the distance between the fine-tuned projection matrix and the pretrained one (\textit{i.e.}, higher $\lambda$).

\section{\method hyperparameters study}
\label{sec:grid_search_hyp}

\begin{figure}[h]
    \centering
    \begin{subfigure}{0.32\linewidth}  
        \centering
        \includegraphics[width=\linewidth]{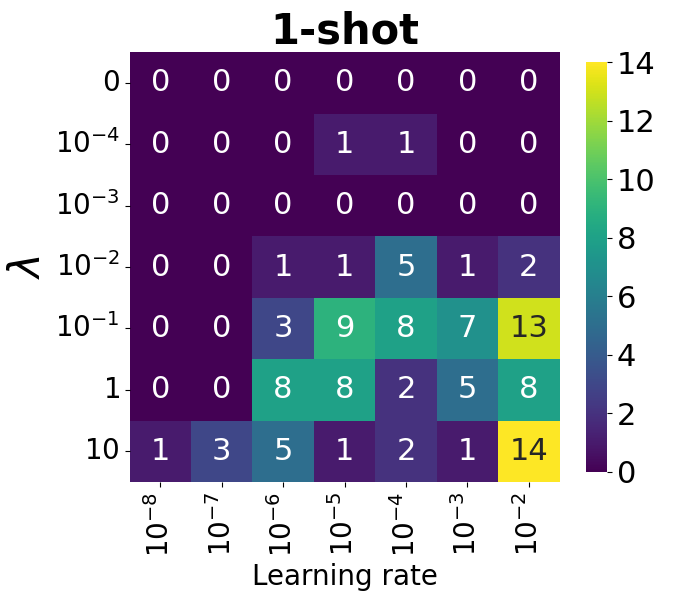}
        \label{fig:subfig1}
    \end{subfigure}%
    \hfill%
    \begin{subfigure}{0.32\linewidth}
        \centering
        \includegraphics[width=\linewidth]{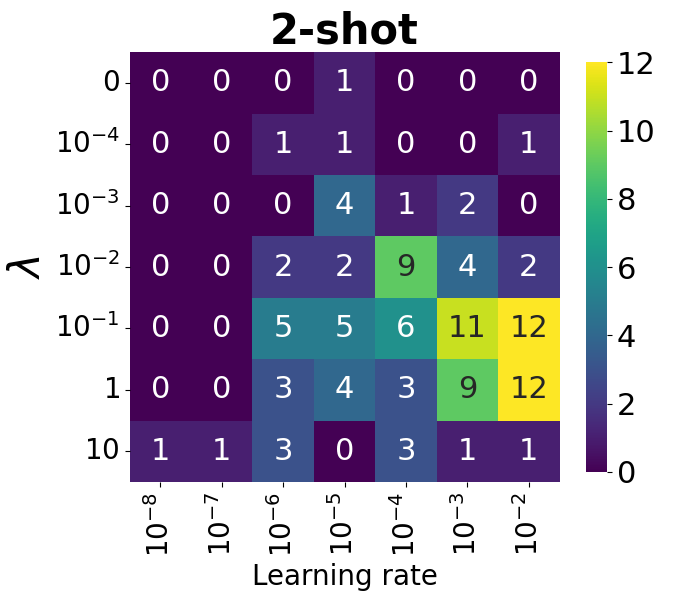}  
        \label{fig:subfig2}
    \end{subfigure}%
    \hfill%
    \begin{subfigure}{0.32\linewidth}
        \centering
        \includegraphics[width=\linewidth]{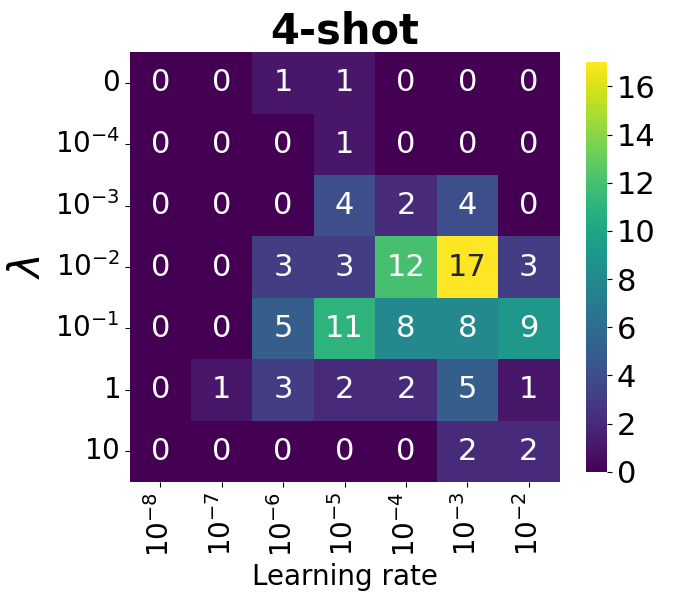}
        \label{fig:subfig3}
    \end{subfigure}\\%
    \begin{subfigure}{0.32\linewidth}
        \centering
        \includegraphics[width=\linewidth]{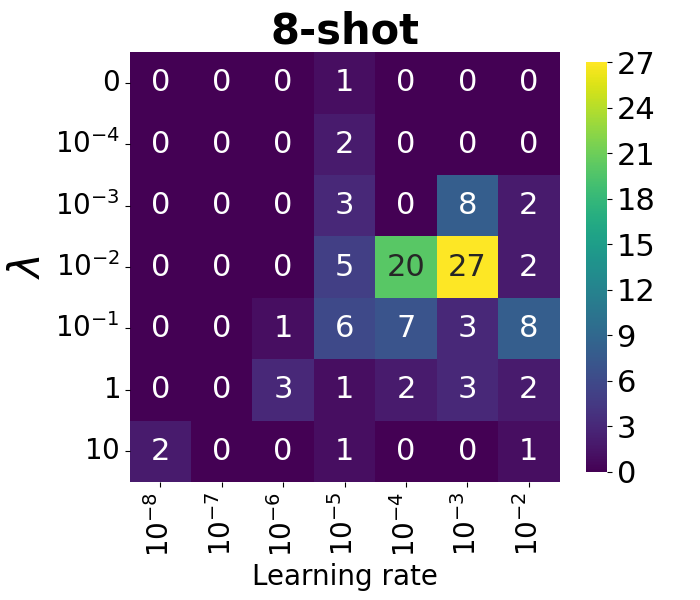}  
        \label{fig:subfig4}
    \end{subfigure}%
    \hspace{0.1\linewidth}%
    \begin{subfigure}{0.32\linewidth}
        \centering
        \includegraphics[width=\linewidth]{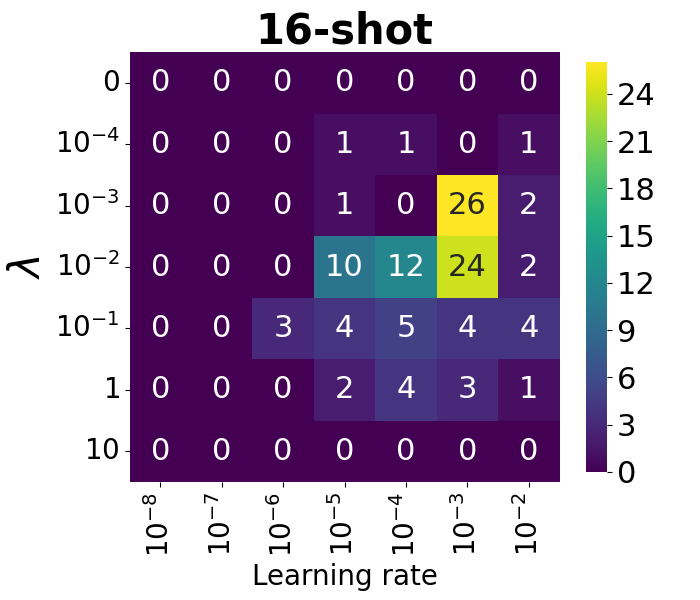} 
        \label{fig:subfig5}
    \end{subfigure}%
    \caption{\textbf{Hyperparameters selected by grid search.} Learning rates and regularization loss weights $\lambda$ found with grid search on the few-shot validation set. The distribution of these hyperparameters are shown for each few-shot setting ($N=1,2,4,8,16$).}
    \label{fig:main_fig_reg}
\end{figure}

\paragraph{Grid search.} \cref{fig:main_fig_reg} shows the distribution of hyperparameters found by grid search on the few-shot validation set~(cf.~\cref{tab:result_sota}). We draw two observations:
\begin{enumerate}
    \item The learning rates span a wide range of values, and high values like $10^{-3}$ and $10^{-2}$ are selected several times, which would cause severe overfitting when no regularization is used (cf.~\cref{tab:results_reg_function} and \cref{fig:results_reg_vs_no_reg_no_search}).
    \item $\lambda=0$ is rarely selected, meaning that based on the few-shot validation set, regularized projection matrices generalize better. 
\end{enumerate}

\label{sec:prolip_sensitivity}
\paragraph{Hyperparameter sensitivity.} \cref{tab:fixed_lr_lambda} complements \cref{fig:results_reg_vs_no_reg_no_search}, where \method is trained for different fixed learning rates, with fixed regularization loss weight $\lambda$. Looking at the values, we make the following observations:
\begin{enumerate}
    \item For low learning rates (\textit{i.e.}, $10^{-5}$, $10^{-6}$), unregularized \method shows good performance for different values of $N$, demonstrating the effectiveness of simply fine-tuning the visual projection matrix. However, the performance drops significantly when the LR {increases.} 
    \item A higher value of $\lambda$ works better for fewer training shots $N$, and vice versa. This effect is increasingly visible when the LR increases. Such observation is expected: with less data we need more regularization as overfitting risk is higher, and this is the base for formulating $\lambda$ as a decreasing function of $N$ (See \cref{tab:results_reg_function}, which shows the detailed numerical results of~\cref{tab:results_wo_validation_plot}).
\end{enumerate}

\begin{table*}
\begin{center}
\resizebox{.58\textwidth}{!}{%
\begin{tabular}{ll c c c c c}
\toprule
\multicolumn{2}{c}{Method} & $N=1$ & 2 & 4 & 8 & 16 \\
\midrule[0.7pt]
\multicolumn{2}{c}{\textcolor{gray}{CLIP (0-shot)}} & \multicolumn{5}{c}{\textcolor{gray}{58.89}} \\
\midrule
\multicolumn{2}{l}{\method (grid search)} & 64.21 & 67.43 & \best{70.58} & \best{73.73} & \best{76.50} \\
\cmidrule{2-7}
\multirow[c]{4}{*}{\method, LR=$10^{-6}$}& $\lambda=1$ & 62.85 & 64.98 & 66.66 & 68.13 & 68.98 \\
&$\lambda=10^{-1}$ & 63.69 & 66.51 & 68.87 & 71.07 & 72.50 \\
& $\lambda=10^{-2}$ & 63.73 & 66.62 & 69.09 & 71.42 & 72.92 \\
&$\lambda=0$ & 63.73 & 66.64 & 69.12 & 71.46 & 72.96 \\
\cmidrule{2-7}
\multirow[c]{4}{*}{\method, LR=$10^{-5}$}
&$\lambda=1$ & 64.28 & 66.59 & 68.30 & 69.67 & 70.49 \\
&$\lambda=10^{-1}$ & \best{64.60} & \second{67.49} & 70.13 & 72.71 & 74.75 \\
&$\lambda=10^{-2}$ & 63.54 & 66.87 & 70.03 & 73.06 & 75.69 \\
&$\lambda=0$ & 62.84 & 66.35 & 69.69 & 72.89 & 75.65 \\
\cmidrule{2-7}
\multirow[c]{4}{*}{\method, LR=$10^{-4}$}
&$\lambda=1$ & 64.40 & 66.86 & 68.82 & 70.37 & 71.36 \\
&$\lambda=10^{-1}$ & \second{64.48} & \best{67.51} & \second{70.37} & 73.08 & 75.25 \\
&$\lambda=10^{-2}$ & 60.45 & 64.73 & 69.04 & 72.85 & 75.80 \\
&$\lambda=0$ & 50.55 & 58.69 & 65.18 & 69.93 & 73.28 \\
\cmidrule{2-7}
\multirow[c]{4}{*}{\method, LR=$10^{-3}$}
&$\lambda=1$ & 64.39 & 66.82 & 68.78 & 70.42 & 71.45  \\
&$\lambda=10^{-1}$ & 64.08 & 67.32 & 70.28 & \second{73.17} & 75.41 \\
&$\lambda=10^{-2}$ & 58.42 & 64.43 & 69.16 & 72.94 & \second{75.99} \\
&$\lambda=0$ & 40.05 & 49.60 & 56.35 & 60.33 & 61.79 \\
\cmidrule{2-7}
\multirow[c]{4}{*}{\method, LR=$10^{-2}$}
&$\lambda=1$ & 64.25 & 66.83 & 68.75 & 70.36 & 71.34 \\
&$\lambda=10^{-1}$ & 63.04 & 67.03 & 70.05 & 72.75 & 74.73 \\
&$\lambda=10^{-2}$ & 53.58 & 61.43 & 67.47 & 71.92 & 75.22 \\
&$\lambda=0$ & 19.98 & 24.12 & 28.03 & 32.42 & 35.62\\
\bottomrule
\end{tabular}
}
\caption{\textbf{\method sensitivity to hyperparameter choice.} Accuracy of \method to the hyperparameters (learning rate LR and regularization weight $\lambda$) for $N\in\{1, 2, 4, 8, 16\}$-shot settings. Each number is an average over 11 datasets, 10 seeds for each.}
\label{tab:fixed_lr_lambda}

\end{center}
\end{table*}

\begin{table*}[h]
\setlength{\fboxsep}{1pt}
\centering
\resizebox{0.63\linewidth}{!}{%
\begin{tabular}{l l c c c c c}
\toprule
\multicolumn{2}{c}{Method} & $N=1$ & 2 & 4 & 8 & 16 \\
\midrule[0.7pt]
\multicolumn{2}{c}{\textcolor{gray}{CLIP (0-shot)}} & \multicolumn{5}{c}{\textcolor{gray}{58.89}} \\
\midrule
\multirow[c]{5}{*}{\methodvalfree, $\lambda=1/N$} 
& LR=$10^{-5}$& 64.28 & 67.07 & 69.68 & 72.57 & 75.20 \\
& LR=$10^{-4}$ & 64.40 & 67.28 & 70.08 & 72.97 & 75.57 \\
&LR=$10^{-3}$ & 64.39 & 67.20 & 70.01 & 73.02 & 75.73\\
&LR=$10^{-2}$ & 64.25 & 67.20 & 69.98 & 72.70 & 75.34 \\
& \cellcolor{shadecolor}Average & \cellcolor{shadecolor}\best{64.33} & \cellcolor{shadecolor}\second{67.19} & \cellcolor{shadecolor}\second{69.94} & \cellcolor{shadecolor}\second{72.82} & \cellcolor{shadecolor}\best{75.46}\\
\midrule
\multirow[c]{5}{*}{\methodvalfree, $\lambda=1/N^2$}
&LR=$10^{-5}$ & 64.28 & 67.32 & 70.22 & 73.10 & 75.68 \\
&LR=$10^{-4}$ & 64.40 & 67.53 & 70.36 & 73.08 & 75.07 \\
&LR=$10^{-3}$ & 64.39 & 67.40 & 70.25 & 73.10 & 75.80 \\
&LR=$10^{-2}$& 64.25 & 67.31 & 70.02 & 72.50 & 74.50 \\
& \cellcolor{shadecolor}Average  & \cellcolor{shadecolor}\best{64.33} & \cellcolor{shadecolor}\best{67.39} & \cellcolor{shadecolor}\best{70.21} & \cellcolor{shadecolor}\best{72.95} & \cellcolor{shadecolor}\second{75.26}\\
\midrule
\multirow[c]{5}{*}{\methodvalfree, $\lambda=0$}
&LR=$10^{-5}$& 62.84 & 66.35 & 69.69 & 72.89 & 75.65 \\
&LR=$10^{-4}$ & 50.55 & 58.69 & 65.18 & 69.93 & 73.28 \\
&LR=$10^{-3}$& 40.05 & 49.60 & 56.35 & 60.33 & 61.79 \\
&LR=$10^{-2}$ & 19.98 & 24.12 & 28.03 & 32.42 & 35.62 \\
& \cellcolor{shadecolor}Average & \cellcolor{shadecolor}43.36 & \cellcolor{shadecolor}49.69 & \cellcolor{shadecolor}54.81 & \cellcolor{shadecolor}58.89 & \cellcolor{shadecolor}61.59\\
\bottomrule
\end{tabular}
}
\caption{\textbf{\methodvalfree with a parametric $\lambda$.} Accuracy (\%) of \methodvalfree with fixed learning rate (LR) and $\lambda$ as a function of $N$. For each $\lambda$ value, we report performance for different LRs and \colorbox{shadecolor}{averaged across LRs}.  Numbers are averages over 11 datasets and 10 seeds. We highlight \best{best} and \second{2nd best} for averages across LRs.}
\label{tab:results_reg_function}
\end{table*}

\section{Details on base-to-new generalization}
\label{sec:app_base_to_new}

\paragraph{Metrics details.} 
Previous works~\cite{zhu2023prompt,khattak2023maple} calculate the \textit{total harmonic mean} over datasets in two different ways.

To extend \cref{tab:base_to_new_res}, in \cref{tab:base_to_new_res_eq8} we report for each architecture both ways of calculating the \textit{total harmonic means}, renaming them ${{\text{H}}_{t1}}$ and ${{\text{H}}_{t2}}$ for disambiguation. It highlights the superiority of our method, regardless of the total harmonic mean used. We also detail the computation below.

\begin{table}[h]
    \centering
    \vspace{-2mm}
    
    \begin{subtable}[t]{0.49\linewidth}
        \centering
        
        \resizebox{1\textwidth}{!}{%
        \begin{tabular}{l cc|c|c}
            \toprule
            & Base & New & ${{\text{H}}_{t1}}$ & ${{\text{H}}_{t2}}$ \\
            \midrule
            \textcolor{gray}{CLIP} & \textcolor{gray}{61.72} & \textcolor{gray}{65.91} & \textcolor{gray}{63.64} & \textcolor{black}{63.75} \\
            \midrule
            CoOp & 71.96 & 61.26 & 65.58 & \textcolor{black}{66.18} \\
            CoCoOp & 72.23 & 60.77 & 65.35 & \textcolor{black}{66.01} \\
            ProGrad & \second{73.29} & \second{65.96} & \second{69.06} & \textcolor{black}{\second{69.43}} \\
            {\methodvalfree} & \best{75.45} & \best{69.43} & \textbf{72.12} & \textcolor{black}{\best{72.31}} \\
            \bottomrule
        \end{tabular}
        }
        \caption{ResNet-50}
    \end{subtable}%
    \hfill
    \begin{subtable}[t]{0.49\linewidth}
        \centering
        \resizebox{1\textwidth}{!}{%
        \begin{tabular}{l cc|c|c}
            \toprule
            & Base & New & ${{\text{H}}_{t1}}$ & ${{\text{H}}_{t2}}$ \\
            \midrule
            \textcolor{gray}{CLIP} & \textcolor{gray}{69.34} & \textcolor{gray}{74.22} & \textcolor{black}{71.59} & \textcolor{gray}{71.70} \\
            \midrule
            CoOp & \second{82.69} & 63.22 & \textcolor{black}{70.83} & 71.66 \\
            CoCoOp & 80.47 & 71.69 & \textcolor{black}{75.44} & 75.83 \\
            MaPLe & 82.28 & \best{75.14} & \textcolor{black}{\second{78.27}} & \second{78.55} \\
            {\methodvalfree} & \best{83.85} & \second{74.78} & \best{78.85} & \textcolor{black}{\best{79.06}} \\
            \bottomrule
        \end{tabular}
        }
        \caption{ViT-B/16}
    \end{subtable}
\caption{\textbf{Base-to-new.} Performance comparison of methods on ResNet-50 and ViT-B/16 architectures across 11 datasets with either $\text{H}_{t1}$ (\eqref{eq:average_hm}) or  $\text{H}_{t2}$ (\eqref{eq:hm_of_average_perfs}).
Numbers highlight the superiority of our method.
}
\label{tab:base_to_new_res_eq8}
\end{table}

In ProGrad~\cite{zhu2023prompt}, the {total harmonic mean} over the $11$ datasets is computed as \textit{the average harmonic means of individual datasets.} This writes:
\begin{equation}
\label{eq:average_hm}
    {{\text{H}}_{t1}} = \frac{1}{11} \sum_{i=1}^{11} \text{HM}_i\,,
\end{equation}
$\text{HM}_{i} = 2 \times \frac{{\text{acc}_\text{b}}_{i} \times {\text{acc}_\text{n}}_{i}}{{\text{acc}_\text{b}}_{i}+{\text{acc}_\text{n}}_{i}}$ being the harmonic mean of dataset $i$. Here, ${\text{acc}_\text{b}}_{i}$ and ${\text{acc}_\text{n}}_i$ denote the accuracy on base and new classes for dataset $i$, respectively.

Instead in MaPLe~\cite{khattak2023maple}, the {total harmonic mean} over the $11$ datasets is calculated as \textit{the harmonic mean of average base and average new classes accuracies}:
\begin{equation}
\label{eq:hm_of_average_perfs}
    {{\text{H}}_{t2}} = 2 \times \frac{\text{acc}_\text{b} \times \text{acc}_\text{n}}{\text{acc}_\text{b}+\text{acc}_\text{n}}\,,
\end{equation}
where $\text{acc}_\text{b}= \frac{1}{11}\sum_{i=1}^{11}{\text{acc}_\text{b}}_i$ and $\text{acc}_\text{n}= \frac{1}{11}\sum_{i=1}^{11}{\text{acc}_\text{n}}_i$.

\paragraph{Per-dataset performance.}
We report in~\cref{tab:app_base2new_resnet} and~\cref{tab:app_base2new_vit} the per-dataset accuracy for base and new classes, as well as the harmonic mean metrics.

\section{{Fine-tuning the text embedding projector}}
\label{sec:text_embed_proj}

\smallskip\noindent\textbf{Can the text embedding projector work?} As discussed in~\cref{sec:method}, CLIP also maps text embeddings to the shared space using a projection matrix. Here, instead of fine-tuning the visual projection matrix $\mW_o$, we fine-tune its textual counterpart $\mW_{ot}$, with the same strategy adopted in \methodvalfree. That is, the visual backbone, including $\mW_o$, is frozen. Only $\mW_{ot}$ is trained with:
\begin{equation}
    L_{\text{ProLIP (text)}} = L(\mW_{ot}) + \lambda \|\mW_{ot} - \mW_{ot}^{(0)}\|_\text{F}^2,
\end{equation}
where $\lambda$ is set to $\frac{1}{N}$.
\cref{tab:text_proj} shows that this variant is also a strong baseline, though underperforming \methodvalfree where the visual embedding projection is fine-tuned. This experiment gives a positive signal on the extendability of our method to other modalities.

\begin{table}[t]
\newcommand{\varh}[1]{} 

\setlength{\fboxsep}{1pt}
\centering
\resizebox{0.75\linewidth}{!}{%
\setlength{\tabcolsep}{0.3em}
\begin{tabular}{l c c c c c}
\toprule
{Method} & $N=1$ & 2 & 4 & 8 & 16 \\
\midrule[0.7pt]
\textcolor{gray}{CLIP (0-shot)} & \multicolumn{5}{c}{\textcolor{gray}{58.89}} \\
\midrule
\methodvalfree (text) & 64.05 & 66.93 & 69.71 & 72.56 & 75.01 \\
\methodvalfree (ours) & \best{64.33} & \best{67.19} & \best{69.94} & \best{72.82} & \best{75.46}\\
\bottomrule
\end{tabular}%
}%
\caption{\textbf{Comparison to fine-tuning the text embedding projection matrix.} We report the classification accuracy (\%) averaged over 11 datasets, 10 seeds, and 4 learning rates $\text{LR} \in \{10^{-5},10^{-4},10^{-3},10^{-2}\}$ where we fine-tune the text projection matrix instead of the visual one, with the same regularization strategy. We call this variant `\methodvalfree (text)' and use $\lambda=1/N$.}
\label{tab:text_proj}
\end{table}

{\cref{tab:text_proj_full_table} complements \cref{tab:text_proj}, showing the performance of this version, coined `\methodvalfree (text)', for different values of LR. We note that this baseline is strong, yet still underperforming \methodvalfree and exhibiting more sensitivity to the choice of LR.}

\begin{table}[h]
\newcommand{\varh}[1]{} 

\setlength{\fboxsep}{1pt}
\centering
\resizebox{0.8\linewidth}{!}{%
\setlength{\tabcolsep}{0.3em}
\begin{tabular}{l l c c c c c}
\toprule
{Method} & LR & $N=1$ & 2 & 4 & 8 & 16 \\
\midrule[0.7pt]
\textcolor{gray}{CLIP (0-shot)} & & \multicolumn{5}{c}{\textcolor{gray}{58.89}} \\
\midrule
\multirow{4}{*}{\methodvalfree (text)} & $10^{-5}$ & 64.25 & 67.10 & 69.91 & 72.82 & 75.34 \\
& $10^{-4}$ & 64.13 & 67.14 & 70.01 & 72.80 & 75.20 \\
& $10^{-3}$ & 63.99 & 66.74 & 69.52 & 72.41 & 75.00 \\
& $10^{-2}$ & 63.81 & 66.72 & 69.39 & 72.21 & 74.51 \\
\bottomrule
\end{tabular}%
}%
\caption{{\textbf{Fine-tuning the text embedding projection matrix.} We report classification accuracy (\%) of `\methodvalfree (text)' averaged over 11 datasets and 10 seeds, using different learning rates (LR).}}
\label{tab:text_proj_full_table}
\end{table}

\section{Details on test-time \method}
\label{sec:app_testtime}

TPT~\citep{shu2022test} learns a single prompt for each test image using an unsupervised loss function. Given a test image $\tI_\text{test}$, the image is augmented $N_\text{views}$ times using a family of random augmentations $\mathcal{A}$. Predictions are made for each view, and the training consists of minimizing the entropy of the averaged probability distribution of these predictions:
\begin{equation}
    \vp^* = \text{argmin}_{\vp} -\sum_{i=1}^{K}\tilde{p}_{\vp}(y_i|\tI_\text{test})\log\tilde{p}_{\vp}(y_i|\tI_\text{test}),
\label{eq:min_ent}
\end{equation}
where
\begin{equation}
    \tilde{p}_{\vp}(y_i|\tI_\text{test}) = \frac{1}{N_\text{views}}\sum_{i=1}^{N_\text{views}}p_{\vp}(y_i|\mathcal{A}_i(\tI_\text{test})).
\label{eq:one_proba}
\end{equation}

In addition, \textit{confidence selection} is used to filter out predictions with high entropy, which are considered as noisy. Self-entropy is computed for each of the $N_\text{views}$; a fixed cutoff percentile $\rho$ keeps only predictions with lower entropy than $\tau$. In Equation~\ref{eq:min_ent}, $\tilde{p}_{\vp}$ becomes:
\begin{equation}
    \tilde{p}_{\vp}(y|\tI_\text{test}) = \frac{1}{\rho N} \sum_{i=1}^{N_\text{views}}1_{\{H(p_i) \leq \tau\}}p_{\vp}(y|\mathcal{A}_i(\tI_\text{test})).
\label{eq:confidence}
\end{equation}

We apply the same framework (\textit{i.e.}, loss function, confidence selection) with the only difference of minimizing Equation~\ref{eq:min_ent} over $\mW_o$ instead of the prompt $\vp$. For a fair comparison, we use the same number of steps for training (\textit{i.e.}, 1 step) and the same value of the cutoff percentile $\rho=0.1$. The learning rate is $10^{-4}$. Note that, measured on ImageNet, \method is $\sim 13$ times faster than TPT, as the latter requires backpropagation trough the whole text encoder, while in our case backpropagation is limited to the visual projection layer and is not applied on the text encoder. We also stress that since we perform only 1 step of training, the regularization loss cannot be used as the first value it takes is $0$ (initially the fine-tuned projection matrix is equal to the pre-trained one).

\begin{table}[t]
\centering
\resizebox{0.80\linewidth}{!}{%
\begin{tabular}{r|c|c|c}
\toprule
Method & \textbf{Cls.} & \textbf{B2N} 
& \textbf{Arch. Agnostic} \\
\midrule
CLIP-LoRA~\cite{zanella2024low} & {\textbf{77.74}}  
& 77.28 & \xmark \\ 
\methodvalfree & 76.26 
& {\textbf{79.06}} & \tick \\ 
\bottomrule
\end{tabular}
}
\caption{\textbf{CLIP-LoRA vs. \methodvalfree.} Few-shot classification (`\textbf{Cls.}') accuracies are averages of 550 runs (11 datasets, 10 seeds, 5 few-shot settings), Base-to-new (`\textbf{B2N}') harmonic means are averages of 110 runs (11 datasets, 10 seeds, N=16-shot setting). For fair comparison, we adopt `a photo of a \{\}.' as template, similarly to CLIP-LoRA.}
\label{tab:clip-lora}
\end{table}

\begin{table}[t]
\centering
\resizebox{0.8\linewidth}{!}{%
\begin{tabular}{c|c|c|c}
\toprule
Arch. & Method & \textbf{Cls.} & \textbf{Params} \\ 
\midrule
\multirow{2}{*}{RN50} & BatchNorm~\cite{wang2021tent} & 71.66 & {\textbf{0.05M}} \\ 
& \methodvalfree & {\textbf{75.46}} & 2.10M \\ 
\midrule
\multirow{2}{*}{ViT-B/16}  & LayerNorm~\cite{zhao2024tuning} & 78.13 & {\textbf{0.07M}} \\ 
& \methodvalfree & {\textbf{81.00}} & 0.39M \\
\bottomrule
\end{tabular}
}
\caption{\textbf{Normalization parameters tuning vs. \methodvalfree}. Few-shot classification (\textbf{`Cls.'}) accuracies are averages of 110 runs (11 datasets \& 10 seeds). The number of samples per class is $N=16$.}
\label{tab:norm-tuning}
\end{table}

\begin{table*}[t]
    \begin{subtable}[t]{.3\linewidth}
    \centering
    \resizebox{0.93\textwidth}{!}{
    \begin{tabular}{l cc|c|c}
            \toprule
            & Base & New & $\text{H}_{t1}$ & $\text{H}_{t2}$ \\
            \midrule
            \textcolor{gray}{CLIP} & \textcolor{gray}{61.72} & \textcolor{gray}{65.91} & \textcolor{gray}{63.64} & \textcolor{gray}{63.75} \\
            \midrule
            CoOp & 71.96 & 61.26 & 65.58 & 66.18 \\
            CoCoOp & 72.23 & 60.77 & 65.35 & 66.01 \\
            ProGrad & \second{73.29} & \second{65.96} & \second{69.06} & \second{69.43} \\
            {\methodvalfree} & \best{75.45} & \best{69.43} & \textbf{72.12} & \best{72.31} \\
            \bottomrule
        \end{tabular}
    }
    \caption{\textbf{Average over 11 datasets}.}
    \end{subtable}
    \hfill
    \begin{subtable}[t]{.2\linewidth}
    \centering
    \resizebox{0.93\textwidth}{!}{
    \begin{tabular}{cc|c}
    \toprule
     Base & New & HM \\
    \midrule
    \textcolor{gray}{64.46} & \textcolor{gray}{59.99} & \textcolor{gray}{62.14}\\
    \midrule
    65.49 &57.70 & 61.35\\
    66.21 &58.01 & 61.84 \\
    66.96 & 60.04 & 63.23\\
    \textbf{67.39} & \textbf{62.24} & \textbf{64.71}\\
    \bottomrule
    \end{tabular}
    }
    \caption{ImageNet}
    \end{subtable}
    \hfill
    \begin{subtable}[t]{.2\linewidth}
    \centering
    \resizebox{0.9\textwidth}{!}{
    \begin{tabular}{cc|c}
    \toprule
     Base & New & HM \\
    \midrule
    \textcolor{gray}{90.90} & \textcolor{gray}{90.72}  & \textcolor{gray}{90.81} \\
    \midrule
    94.38 & 87.48 & 90.80\\
    94.43 & 87.81 & 91.00  \\
    94.47 & 90.84  & 92.46 \\
    \textbf{95.39} & \textbf{91.15} & \textbf{93.22} \\
    \bottomrule
    \end{tabular}
    }
    \caption{Caltech101}
    \end{subtable}
    \hfill
    \begin{subtable}[t]{.2\linewidth}
    \centering
    \resizebox{0.9\textwidth}{!}{
    \begin{tabular}{cc|c}
    \toprule
    Base & New & HM \\
    \midrule
    \textcolor{gray}{85.86}  & \textcolor{gray}{93.85} & \textcolor{gray}{89.68}  \\
    \midrule
    90.31  & 94.03 & 92.13 \\
    89.07  & 91.00 & 90.02 \\
    \textbf{91.78}  & \textbf{94.86} & \textbf{93.29} \\
    90.86 & 93.13 & 91.98 \\
    \bottomrule
    \end{tabular}
    }
    \caption{OxfordPets}
    \end{subtable}
    \hfill
    \begin{subtable}[t]{.3\textwidth}
    \centering
    \resizebox{0.9\textwidth}{!}{
    \begin{tabular}{l cc|c}
    \toprule
    & Base & New & HM \\
    \midrule
    \textcolor{gray}{CLIP} &  \textcolor{gray}{55.55} & \textcolor{gray}{66.35} & \textcolor{gray}{60.47} \\
    \midrule
    CoOp & 61.77  & 62.51 & 62.14\\
    CoCoOp & 61.68& 59.98 &60.82\\
    ProGrad & 63.01 & 64.32 & 63.66 \\
    \methodvalfree & \textbf{64.61} & \textbf{65.93} & \textbf{65.26} \\
    \bottomrule
    \end{tabular}
    }
    \caption{StanfordCars}
    \end{subtable}
    \hfill
    \begin{subtable}[t]{.2\textwidth}
    \centering
    \resizebox{0.9\textwidth}{!}{
    \begin{tabular}{ cc|c}
    \toprule
     Base & New & HM \\
    \midrule
    \textcolor{gray}{64.10}  & \textcolor{gray}{70.92}  & \textcolor{gray}{67.34} \\
    \midrule
    89.33  & 62.77  & 73.73 \\
    88.07  & 66.26 & 75.62 \\
    88.19 & 69.38 & 77.66 \\
    \textbf{89.42} & \textbf{72.34} & \textbf{79.98}\\
    \bottomrule
    \end{tabular}
    }
    \caption{Flowers102}
    \end{subtable}
    \hfill
    \begin{subtable}[t]{.2\textwidth}
    \centering
    \resizebox{0.9\textwidth}{!}{
    \begin{tabular}{cc|c}
    \toprule
    Base & New & HM \\
    \midrule
    \textcolor{gray}{81.48} & \textcolor{gray}{82.15} & \textcolor{gray}{81.81} \\
    \midrule
    80.40 & 81.09 & 80.74 \\
    79.77 & 77.68 & 78.71 \\
    \textbf{83.10}  & 83.57 & 83.33 \\
    82.39 & \textbf{84.47} & \textbf{83.42} \\
    \bottomrule
    \end{tabular}
    }
    \caption{Food101}
    \end{subtable}
    \hfill
    \begin{subtable}[t]{.2\textwidth}
    \centering
    \resizebox{0.9\textwidth}{!}{
    \begin{tabular}{ cc|c}
    \toprule
     Base & New & HM \\
    \midrule
    \textcolor{gray}{17.89} & \textcolor{gray}{25.13} & \textcolor{gray}{20.90} \\
    \midrule
    22.53 & 20.40 & 21.41\\
    22.73 & 19.40 & 20.93 \\
    22.77 & 24.24 & 23.48 \\
    \textbf{26.67} & \textbf{26.92} & \textbf{26.79} \\
    \bottomrule
    \end{tabular}
    }
    \caption{FGVCAircraft}
    \end{subtable}
    \hfill
    \begin{subtable}[t]{.3\textwidth}
    \centering
    \resizebox{0.9\textwidth}{!}{
    \begin{tabular}{l cc|c}
    \toprule
    & Base & New & HM \\
    \midrule
    \textcolor{gray}{CLIP} & \textcolor{gray}{66.45} & \textcolor{gray}{70.17} & \textcolor{gray}{68.26} \\
    \midrule
    CoOp & 71.48 & 65.57 & 68.40\\
    CoCoOp & 71.88 & 67.10 & 69.41 \\
    ProGrad & 73.71  & 69.78 & 71.69 \\
    \methodvalfree & \textbf{75.20} & \textbf{72.69} & \textbf{73.92} \\
    \bottomrule
    \end{tabular}
    }
    \caption{SUN397}
    \end{subtable}
    \hfill
    \begin{subtable}[t]{.2\textwidth}
    \centering
    \resizebox{0.9\textwidth}{!}{
    \begin{tabular}{ cc|c}
    \toprule
    Base & New & HM \\
    \midrule
     \textcolor{gray}{49.31}  & \textcolor{gray}{54.35} & \textcolor{gray}{51.71} \\
     \midrule
     67.71  & 43.92 & 53.28 \\
     63.54  & 40.78 & 49.68\\
     66.90  & 53.06  & 59.18 \\
     \textbf{71.00} & \textbf{57.09} & \textbf{63.29} \\
    \bottomrule
    \end{tabular}
    }
    \caption{DTD}
    \end{subtable}
    \hfill
    \begin{subtable}[t]{.2\textwidth}
    \centering
    \resizebox{0.9\textwidth}{!}{
    \begin{tabular}{ cc|c}
    \toprule
     Base & New & HM \\
    \midrule
     \textcolor{gray}{39.26}  & \textcolor{gray}{43.62} & \textcolor{gray}{41.33} \\
     \midrule
     73.53 & 40.19 & 51.97\\
     83.63  & 40.95  & 54.98 \\
     79.67  & 49.99 & 61.43 \\
    \textbf{88.16} & \textbf{66.69} & \textbf{75.94}\\
    \bottomrule
    \end{tabular}
    }
    \caption{EuroSAT}
    \end{subtable}
    \hfill
    \begin{subtable}[t]{.2\textwidth}
    \centering
    \resizebox{0.9\textwidth}{!}{
    \begin{tabular}{cc|c}
    \toprule
    Base & New & HM \\
    \midrule
     \textcolor{gray}{63.70} & \textcolor{gray}{67.71} & \textcolor{gray}{65.64} \\
     \midrule
     74.59  &58.23 & 65.40\\
     73.51  & 59.55 & 65.80  \\
     75.66   & 65.52  & 70.23 \\
    \textbf{78.89} & \textbf{71.13} & \textbf{74.81}\\
    \bottomrule
    \end{tabular}
    }
    \caption{UCF101}
    \end{subtable}
    ~
    \caption{\textbf{Base-to-new generalization with ResNet-50.} Per-dataset base, new, and harmonic mean accuracy of \methodvalfree with $N=4$ (except `CLIP' which is zero-shot); cf.~\cref{tab:base_to_new_res}(a).}
    \label{tab:app_base2new_resnet}
\end{table*}

\begin{table*}[t]
    \begin{subtable}[t]{.3\linewidth}
    \centering
    \resizebox{0.93\textwidth}{!}{
    \begin{tabular}{l cc|c|c}
            \toprule
            & Base & New & $\text{H}_{t1}$ & $\text{H}_{t2}$ \\
            \midrule
            \textcolor{gray}{CLIP} & \textcolor{gray}{69.34} & \textcolor{gray}{74.22} & \textcolor{gray}{71.59} & \textcolor{gray}{71.70} \\
            \midrule
            CoOp & \second{82.69} & 63.22 & 70.83 & 71.66 \\
            CoCoOp & 80.47 & 71.69 & 75.44 & 75.83 \\
            MaPLe & 82.28 & \best{75.14} & \second{78.27} & \second{78.55} \\
            {\methodvalfree} & \best{83.85} & \second{74.78} & \best{78.85} & \best{79.06} \\
            \bottomrule
        \end{tabular}
    }
    \caption{\textbf{Average over 11 datasets}.}
    \end{subtable}
    \hfill
    \begin{subtable}[t]{.2\linewidth}
    \centering
    \resizebox{0.93\textwidth}{!}{
    \begin{tabular}{cc|c}
    \toprule
     Base & New & HM \\
    \midrule
    \textcolor{gray}{72.43} & \textcolor{gray}{68.14} & \textcolor{gray}{70.22}\\
    \midrule
    76.47 & 67.88 & 71.92\\
    75.98 & 70.43 & 73.10 \\
    \textbf{76.66} & \textbf{70.54} & \textbf{73.47}\\
    76.56 & 68.63 & 72.38\\
    \bottomrule
    \end{tabular}
    }
    \caption{ImageNet}
    \end{subtable}
    \hfill
    \begin{subtable}[t]{.2\linewidth}
    \centering
    \resizebox{0.9\textwidth}{!}{
    \begin{tabular}{cc|c}
    \toprule
     Base & New & HM \\
    \midrule
    \textcolor{gray}{96.84} & \textcolor{gray}{94.00} & \textcolor{gray}{95.40} \\
    \midrule
    98.00 & 89.81 & 93.73\\
    97.96 & 93.81 & 95.84  \\
    97.74 & 94.36 & 96.02 \\
    \textbf{98.55} & \textbf{94.39} & \textbf{96.43} \\
    \bottomrule
    \end{tabular}
    }
    \caption{Caltech101}
    \end{subtable}
    \hfill
    \begin{subtable}[t]{.2\linewidth}
    \centering
    \resizebox{0.9\textwidth}{!}{
    \begin{tabular}{cc|c}
    \toprule
    Base & New & HM \\
    \midrule
    \textcolor{gray}{91.17} & \textcolor{gray}{97.26} & \textcolor{gray}{94.12}  \\
    \midrule
    93.67 & 95.29 & 94.47 \\
    95.20 & 97.69 & 96.43 \\
    \textbf{95.43} & \textbf{97.76} & \textbf{96.58} \\
    94.96 & 96.64 & 95.79 \\
    \bottomrule
    \end{tabular}
    }
    \caption{OxfordPets}
    \end{subtable}
    \hfill
    \begin{subtable}[t]{.3\textwidth}
    \centering
    \resizebox{0.9\textwidth}{!}{
    \begin{tabular}{l cc|c}
    \toprule
    & Base & New & HM \\
    \midrule
    \textcolor{gray}{CLIP} &  \textcolor{gray}{63.37} & \textcolor{gray}{74.89} & \textcolor{gray}{68.65} \\
    \midrule
    CoOp & 78.12 & 60.40 & 68.13\\
    CoCoOp & 70.49 & 73.59 & 72.01\\
    MaPLe & 72.94 & \textbf{74.00} & 73.47 \\
    \methodvalfree & \textbf{79.30} & 70.64 & \textbf{74.72}\\
    \bottomrule
    \end{tabular}
    }
    \caption{StanfordCars}
    \end{subtable}
    \hfill
    \begin{subtable}[t]{.2\textwidth}
    \centering
    \resizebox{0.9\textwidth}{!}{
    \begin{tabular}{ cc|c}
    \toprule
     Base & New & HM \\
    \midrule
    \textcolor{gray}{72.08} & \textcolor{gray}{77.80} & \textcolor{gray}{74.83} \\
    \midrule
    97.60 & 59.67 & 74.06 \\
    94.87 & 71.75 & 81.71 \\
    95.92 & 72.46 & 82.56 \\
    \textbf{96.14} & \textbf{74.09} & \textbf{83.69} \\
    \bottomrule
    \end{tabular}
    }
    \caption{Flowers102}
    \end{subtable}
    \hfill
    \begin{subtable}[t]{.2\textwidth}
    \centering
    \resizebox{0.9\textwidth}{!}{
    \begin{tabular}{cc|c}
    \toprule
    Base & New & HM \\
    \midrule
    \textcolor{gray}{90.10} & \textcolor{gray}{91.22} & \textcolor{gray}{90.66} \\
    \midrule
    88.33 & 82.26 & 85.19 \\
    90.70 & 91.29 & 90.99 \\
    \textbf{90.71} & \textbf{92.05} & \textbf{91.38} \\
    90.25 & 90.58 & 90.41 \\
    \bottomrule
    \end{tabular}
    }
    \caption{Food101}
    \end{subtable}
    \hfill
    \begin{subtable}[t]{.2\textwidth}
    \centering
    \resizebox{0.9\textwidth}{!}{
    \begin{tabular}{ cc|c}
    \toprule
     Base & New & HM \\
    \midrule
    \textcolor{gray}{27.19} & \textcolor{gray}{36.29} & \textcolor{gray}{31.09} \\
    \midrule
    40.44 & 22.30 & 28.75\\
    33.41 & 23.71 & 27.74 \\
    37.44 & \textbf{35.61} & 36.50 \\
    \textbf{44.01} & 33.94 & \textbf{38.32} \\
    \bottomrule
    \end{tabular}
    }
    \caption{FGVCAircraft}
    \end{subtable}
    \hfill
    \begin{subtable}[t]{.3\textwidth}
    \centering
    \resizebox{0.9\textwidth}{!}{
    \begin{tabular}{l cc|c}
    \toprule
    & Base & New & HM \\
    \midrule
    \textcolor{gray}{CLIP} & \textcolor{gray}{69.36} & \textcolor{gray}{75.35} & \textcolor{gray}{72.23} \\
    \midrule
    CoOp & 80.60 & 65.89 & 72.51\\
    CoCoOp & 79.74 & 76.86 & 78.27 \\
    MaPLe & 80.82 & \textbf{78.70} & \textbf{79.75} \\
    \methodvalfree & \textbf{82.22} & 77.29 & 79.68\\
    \bottomrule
    \end{tabular}
    }
    \caption{SUN397}
    \end{subtable}
    \hfill
    \begin{subtable}[t]{.2\textwidth}
    \centering
    \resizebox{0.9\textwidth}{!}{
    \begin{tabular}{ cc|c}
    \toprule
    Base & New & HM \\
    \midrule
     \textcolor{gray}{53.24} & \textcolor{gray}{59.90} & \textcolor{gray}{56.37} \\
     \midrule
     79.44 & 41.18 & 54.24 \\
     77.01 & 56.00 & 64.85\\
     80.36 & 59.18 & 68.16 \\
      \textbf{81.44} & \textbf{60.23} & \textbf{69.25} \\
    \bottomrule
    \end{tabular}
    }
    \caption{DTD}
    \end{subtable}
    \hfill
    \begin{subtable}[t]{.2\textwidth}
    \centering
    \resizebox{0.9\textwidth}{!}{
    \begin{tabular}{ cc|c}
    \toprule
     Base & New & HM \\
    \midrule
     \textcolor{gray}{56.48} & \textcolor{gray}{64.05} & \textcolor{gray}{60.03} \\
     \midrule
     92.19 & 54.74 & 68.69 \\
     87.49 & 60.04 & 71.21 \\
     \textbf{94.07} & 73.23 & 82.35 \\
     92.35 & \textbf{77.23} & \textbf{84.12}\\
    \bottomrule
    \end{tabular}
    }
    \caption{EuroSAT}
    \end{subtable}
    \hfill
    \begin{subtable}[t]{.2\textwidth}
    \centering
    \resizebox{0.9\textwidth}{!}{
    \begin{tabular}{cc|c}
    \toprule
    Base & New & HM \\
    \midrule
     \textcolor{gray}{70.53} & \textcolor{gray}{77.50} & \textcolor{gray}{73.85} \\
     \midrule
     84.69 & 56.05 & 67.46\\
     82.33 & 73.45 & 77.64  \\
     83.00 & 78.66 & 80.77 \\
      \textbf{86.57} & \textbf{78.91} & \textbf{82.56}\\
    \bottomrule
    \end{tabular}
    }
    \caption{UCF101}
    \end{subtable}
    \caption{\textbf{Base-to-new generalization with ViT-B/16.} Per-dataset base, new, and harmonic mean accuracy of \methodvalfree with $N=16$ (except `CLIP' which is zero-shot); cf.~\cref{tab:base_to_new_res}(b).}
    \label{tab:app_base2new_vit}
\end{table*}

\section{Comparison to architecture-specific baselines}
\label{sec:arch_spec}

\textbf{Comparison to low-rank adaptation (LoRA).}~Zanella \etal~\cite{zanella2024low} recently showed that applying low-rank adaptation (LoRA)~\cite{hu2022lora} to CLIP is a comptetive baseline to adapters and prompt learning. They apply LoRA on query, key and value matrices of the ViT, and show strong performance on the few-shot classification setting. 
We compare \methodvalfree to CLIP-LoRA~\cite{zanella2024low} for ViT-B/16 on few-shot classification and base-to-new generalization. Few-shot classification results are averaged across $5$ settings (\textit{i.e.}, N=\{1,2,4,8,16\}), $11$ datasets and $10$ seeds, while base-to-new generalization results are averaged across $11$ datasets and $10$ seeds for $N=16$. For a fair comparison, we use the template `a photo of a \{\}' for the class names, similarly to CLIP-LoRA. 
The results are shown in~\cref{tab:clip-lora}. Slower than \method, CLIP-LoRA works well on classification~\cite{zanella2024low} but compromises base-to-new generalization (B2N)~\cite{farina2025rethinking} and is specific to ViT.

\noindent \textbf{Comparison to normalization techniques.} We compare here \methodvalfree to methods that fine-tune only affine normalization parameters for BatchNorm~\cite{wang2021tent} in CNN and LayerNorm~\cite{zhao2024tuning} in ViT. 
Re-purposing these methods to few-shot settings requires full backpropagation and mini-batch training. Despite having more parameters, \methodvalfree is much faster and largely outperforms both baselines. Results are shown in~\cref{tab:norm-tuning}. 

\section{{Further analysis and discussion}}
\label{sec:further_analysis}

\paragraph{Complementarity to other methods.} We showed in \cref{tab:compl} that \method is complementary to other methods that learn different components for few-shot adaptation. Recently, ~\citet{tang2024amu} proposed interpreting CLIP few-shot adaptation methods from a unified perspective of logit bias. That is, every method learns a bias on top of the zero-shot CLIP logits. We detail here the bias learned by each the two methods \method was shown to be complementary to: TaskRes and Tip-Adapter-F, as well as the bias learned by \method.
TaskRes learns an element-wise adapter on top of $\vt$, the text-based frozen classifier. It writes:
\begin{equation}
    \text{Logits}_{\text{TaskRes}} = \vv^\intercal (\vt + \alpha {\vr}) =  \underbrace{\vv^\intercal \vt}_{\text{zero-shot logits}}+ \alpha \vv^\intercal {\vr}.
\label{eq:task_res}
\end{equation}

The bias learned by TaskRes is thus a new linear probe trained on top of frozen visual features $\vv$.

Tip-Adapter-F builds a cache model from the training features $\mF_\text{train}$ and their labels $\mL_\text{train}$. It writes:
\begin{equation}
    \text{Logits}_{\text{Tip-Adapter-F}} = \underbrace{\vv^\intercal \vt}_{\text{zero-shot logits}} + \alpha \phi(\vv^\intercal {\mF_\text{train}^\intercal})\mL_\text{train}. 
\end{equation}

$\mF_\text{train}$ is fine-tuned, thus the bias is based on intra-modal similarity measures (\textit{i.e.}, similarities in the visual space).

For \method, we fine-tune the projection matrix $\mW_o$. Omitting $b_o$ for simplicity, the logits can be written as:
\begin{equation}
    \text{Logits}_{\text{\method}} = \vx_o^\intercal \mW_o^\intercal \vt = \underbrace{\vx_o^\intercal {\mW_o^{(0)}}^{\intercal} \vt}_{\text{zero-shot logits}} + \vx_o^\intercal {\mB}^{\intercal} \vt.
\end{equation}

That is, fine-tuning $\mW_o$ is equivalent to learning a matrix $\mB$, initialized with $\mathbf{0}_{D \times D_o}$. Thus, the bias learned by \method is a linear combination of the pre-projected features, trained to match the fixed text-based probe $\vt$.
In short, each of the three methods learn a different bias, and we hypothesis that the results of \cref{tab:compl} reflect that these biases contain orthogonal knowledge learned during few-shot adaptation.

It is worth noting that we fixed the LR to $10^{-4}$ for all the datasets in these experiments. While the complementarity was shown for fixed hyperparameters across all datasets, ($\alpha=\beta=1$ for Tip-Adapter-F and $\alpha=0.1$ for TaskRes), increasing the LR to $10^{-2}$ lead to overfitting since the biases of TaskRes and Tip-Adapter-F are not regularized, which highlight again the advantage of \method in stability across LRs.

\paragraph{Revisiting CLIP-Adapter~\cite{gao2024clip} with \method{}'s principles.}
\cref{tab:linear_adapter} reports detailed results of CLIP-Adapter when varying its residual weight ($\alpha$) and the learning rate (LR).
Not only are the averaged results significantly worse than those of the Regularized Linear Adapter (RLA) variant and \methodvalfree, but CLIP-Adapter also exhibits high variance, especially in low-shot settings.
Incorporating the \method's principles, RLA consistently improves performance while being much more stable. 
Our~\methodvalfree still achieves the best results.

\begin{table}[t]
\begin{center}
\resizebox{1.0\linewidth}{!}{%
\begin{tabular}{l l c c c c c c}
\toprule
\multicolumn{2}{c}{Method} & $N=1$ & 2 & 4 & 8 & 16 & Average\\
\midrule[0.7pt]
\multirow[c]{5}{*}{$\alpha=0$} & LR=$10^{-5}$& 17.92 & 30.80 & 44.39 & 55.41 & 63.02 & 42.31 \\
& LR=$10^{-4}$& 39.17 & 50.45 & 59.91 & 66.78 & 71.74 & 57.61 \\
& LR=$10^{-3}$& 41.79 & 51.78 & 60.04 & 66.41 & 71.14 & 58.23 \\
& LR=$10^{-2}$& 39.36 & 45.45 & 49.47 & 52.34 & 53.28 & 47.98 \\
& \cellcolor{shadecolor}Average & \cellcolor{shadecolor}34.56 & \cellcolor{shadecolor}44.62 & \cellcolor{shadecolor}53.45 & \cellcolor{shadecolor}60.24 & \cellcolor{shadecolor}64.80 & \cellcolor{shadecolor}51.53 \\

\midrule
\multirow[c]{5}{*}{$\alpha=0.1$} & LR=$10^{-5}$& 57.65 & 62.21 & 66.46 & 70.30 & 73.12 & 65.95 \\
& LR=$10^{-4}$& 57.40 & 62.26 & 66.84 & 70.97 & 74.39 & 66.37 \\
& LR=$10^{-3}$& 44.81 & 53.50 & 61.25 & 67.13 & 71.49 & 59.64 \\
& LR=$10^{-2}$& 38.42 & 44.90 & 49.05 & 51.76 & 53.07 & 47.44 \\
& \cellcolor{shadecolor}Average & \cellcolor{shadecolor}49.57 & \cellcolor{shadecolor}55.72 & \cellcolor{shadecolor}60.90 & \cellcolor{shadecolor}65.04 & \cellcolor{shadecolor}68.02 & \cellcolor{shadecolor}59.85 \\

\midrule
\multirow[c]{5}{*}{$\alpha=0.3$} & LR=$10^{-5}$& 63.39 & 66.62 & 69.39 & 71.88 & 73.69 & 68.99 \\
& LR=$10^{-4}$& 60.26 & 64.28 & 68.23 & 71.96 & 75.12 & 67.97 \\
& LR=$10^{-3}$& 50.77 & 58.05 & 63.86 & 68.77 & 72.74 & 62.84 \\
& LR=$10^{-2}$& 37.55 & 44.37 & 49.22 & 52.07 & 54.69 & 47.58 \\
& \cellcolor{shadecolor}Average & \cellcolor{shadecolor}52.99 & \cellcolor{shadecolor}58.33 & \cellcolor{shadecolor}62.68 & \cellcolor{shadecolor}66.17 & \cellcolor{shadecolor}69.06 & \cellcolor{shadecolor}61.85 \\
\midrule
\multirow[c]{5}{*}{$\alpha=0.5$} & LR=$10^{-5}$& 63.79 & 66.61 & 68.72 & 70.40 & 71.48 & 68.20 \\
& LR=$10^{-4}$& 60.78 & 64.75 & 68.48 & 72.03 & 74.94 & 68.20 \\
& LR=$10^{-3}$& 55.47 & 60.73 & 65.62 & 69.90 & 73.54 & 65.05 \\
& LR=$10^{-2}$& 35.07 & 41.79 & 45.50 & 47.73 & 50.72 & 44.16 \\
& \cellcolor{shadecolor}Average & \cellcolor{shadecolor}53.78 & \cellcolor{shadecolor}58.47 & \cellcolor{shadecolor}62.08 & \cellcolor{shadecolor}65.02 & \cellcolor{shadecolor}67.67 & \cellcolor{shadecolor}61.40 \\
\midrule
\multirow[c]{5}{*}{$\alpha=0.7$} & LR=$10^{-5}$& 63.18 & 64.96 & 66.01 & 66.57 & 66.88 & 65.52 \\
& LR=$10^{-4}$& 61.32 & 65.19 & 68.69 & 71.75 & 74.02 & 68.19 \\
& LR=$10^{-3}$& 56.98 & 61.63 & 66.15 & 70.41 & 74.05 & 65.84 \\
& LR=$10^{-2}$& 36.73 & 41.80 & 43.91 & 46.28 & 47.87 & 43.32 \\
& \cellcolor{shadecolor}Average & \cellcolor{shadecolor}54.55 & \cellcolor{shadecolor}58.40 & \cellcolor{shadecolor}61.19 & \cellcolor{shadecolor}63.75 & \cellcolor{shadecolor}65.71 & \cellcolor{shadecolor}60.72 \\
\midrule
\multirow[c]{5}{*}{$\alpha=0.9$} & LR=$10^{-5}$& 60.74 & 60.99 & 61.04 & 61.12 & 61.13 & 61.00 \\
& LR=$10^{-4}$& 62.42 & 65.40 & 67.38 & 68.69 & 69.40 & 66.66 \\
& LR=$10^{-3}$& 58.55 & 63.13 & 67.47 & 71.23 & 74.14 & 66.90 \\
& LR=$10^{-2}$& 51.42 & 56.48 & 61.59 & 66.71 & 70.91 & 61.42 \\
& \cellcolor{shadecolor}Average & \cellcolor{shadecolor}58.28 & \cellcolor{shadecolor}61.50 & \cellcolor{shadecolor}64.37 & \cellcolor{shadecolor}66.94 & \cellcolor{shadecolor}68.90 & \cellcolor{shadecolor}64.00 \\
\midrule
\multirow[c]{5}{*}{RLA, $\lambda=1/N$} & LR=$10^{-5}$& 63.23 & 65.56 & 68.08 & 70.90 & 73.44 & 68.24 \\
& LR=$10^{-4}$& 63.41 & 65.80 & 68.27 & 70.91 & 73.35 & 68.35 \\
& LR=$10^{-3}$& 63.38 & 65.82 & 68.30 & 70.99 & 73.55 & 68.41 \\
& LR=$10^{-2}$& 63.35 & 65.78 & 68.29 & 70.99 & 73.57 & 68.40 \\
& \cellcolor{shadecolor}Average & \cellcolor{shadecolor}\second{63.34} & \cellcolor{shadecolor}\second{65.74} & \cellcolor{shadecolor}\second{68.24} & \cellcolor{shadecolor}\second{70.95} & \cellcolor{shadecolor}\second{73.48} & \cellcolor{shadecolor}\second{68.35} \\
\midrule
\multirow[c]{5}{*}{\methodvalfree, $\lambda=1/N$}  & LR=$10^{-5}$& 64.28 & 67.07 & 69.68 & 72.57 & 75.20 & 69.76\\
& LR=$10^{-4}$& 64.40 & 67.28 & 70.08 & 72.97 & 75.57 & 70.06\\
& LR=$10^{-3}$& 64.39 & 67.20 & 70.01 & 73.02 & 75.73 & 70.07\\
& LR=$10^{-2}$& 64.25 & 67.20 & 69.98 & 72.70 & 75.34 & 69.89\\
& \cellcolor{shadecolor}Average & \cellcolor{shadecolor}\best{64.33} & \cellcolor{shadecolor}\best{67.19} & \cellcolor{shadecolor}\best{69.94} & \cellcolor{shadecolor}\best{72.82} & \cellcolor{shadecolor}\best{75.46} & \cellcolor{shadecolor}\best{69.95} \\
\bottomrule
\end{tabular}
}
\caption{\textbf{Improving CLIP-Adapter with ProLIP's principles} results in the Regularized Linear Adapter (RLA) variant. We report classification accuracy (\%) averaged over 11 datasets, 10 seeds, and 4 learning rates $\text{LR} {\in} \{10^{-5},10^{-4},10^{-3},10^{-2}\}$ for CLIP-Adapter with different $\alpha$ values, \methodvalfree and RLA with $\lambda=1/N$.}
\label{tab:linear_adapter}
\end{center}
\end{table}

\begin{figure}[t]
    \centering
    \includegraphics[width=0.52\linewidth]{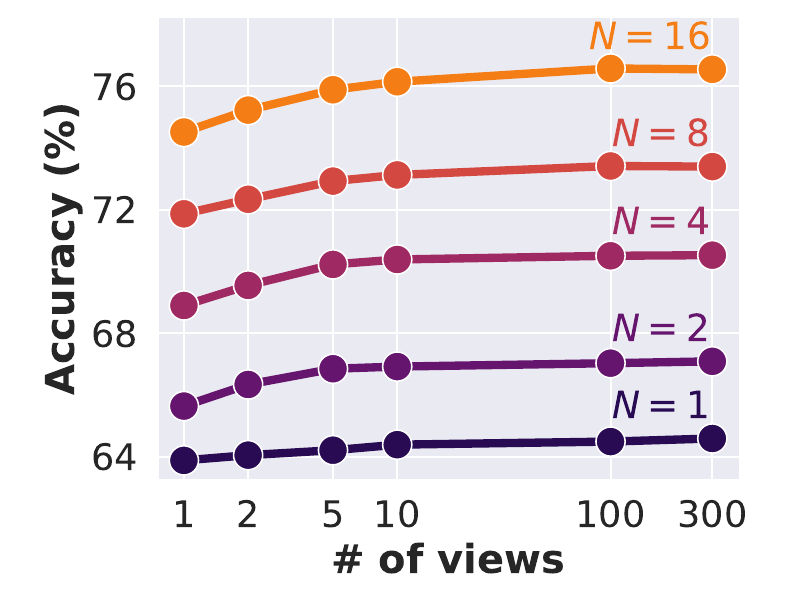}
    \caption{\textbf{Effect of augmented views.} Ablation of ProLIP using varying number of views and shots.}
    \label{fig:view_ablation}
    \hfill%
\end{figure}

\smallskip\noindent\textbf{Number of augmented views.} Following the literature~\cite{zhang2022tip,huang2024lp++}, we apply \texttt{RandomResizedCrop} and \texttt{RandomHorizontalFlip} augmentations during training. As mentionned earlier, ProLIP can be applied on pre-computed visual embeddings (before the projection layer). 
We ablate the number of views in which the features are saved.~\cref{fig:view_ablation} shows that average accuracy over 11 datasets increases with more views. Interestingly, $\sim10$ views are sufficient to get results close to those with $300$ views. In contrast, Lin et al.~\cite{lin2023multimodality} showed that the gain saturates after more than two views for their cross-modal linear probe.

\begin{figure}[t]
    \centering
        \begin{subfigure}[m]{0.49\linewidth}
            \centering
            \includegraphics[width=\linewidth]{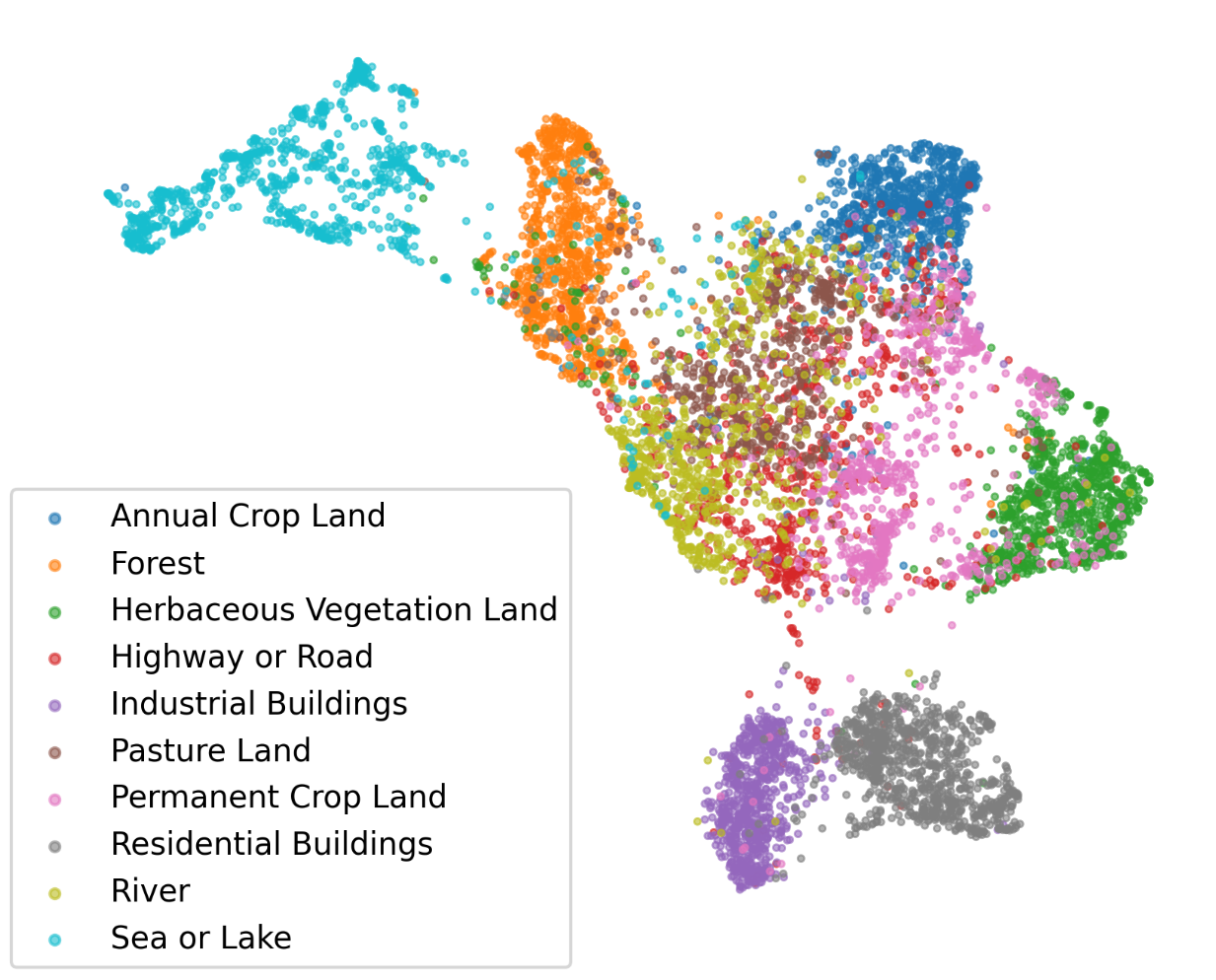}
            \caption{Zero-shot CLIP}
            \label{fig:figure1}
        \end{subfigure}%
        \hfill%
        \begin{subfigure}[m]{0.49\linewidth}
            \centering
            \includegraphics[width=\linewidth]{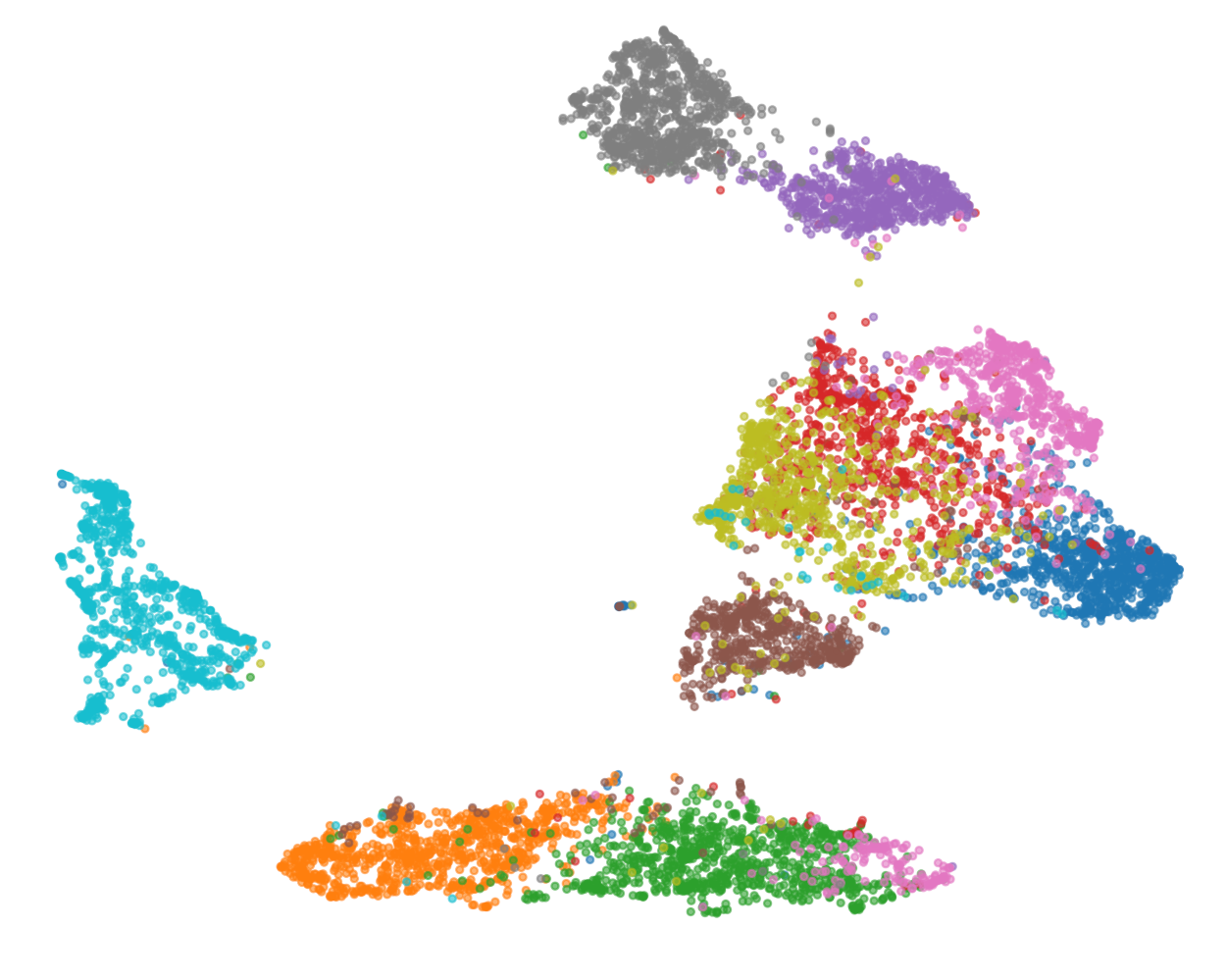}
            \caption{\method}
            \label{fig:figure2}
        \end{subfigure}%
    \caption{\textbf{Ablation and UMAP Visualization.} \subref{fig:figure1} and \subref{fig:figure2}~UMAP of Zero-shot CLIP vs. \method on EuroSAT, showing that some classes (\textit{e.g.}, `Pasture Land', `Permanent Crop Land', `Sea or Lake', etc.) are better clustered with our method.
    }
    \label{fig:combined_plot}
\end{figure}

\smallskip\noindent\textbf{Visualization.}
\label{sec:visualization}
We use UMAP to visualize EuroSAT test set feature manifolds, before and after 16-shot training (\textit{i.e.}, zero-shot \textit{vs.} \method). The results are illustrated in~\cref{fig:figure1,fig:figure2}. We observe that the features are generally better clustered for \method. Confusing categories like \textit{Highway or Road, Permanent Crop Land and Pasture Land} exhibit remarkably better separation for our few-shot adapted model compared to zero-shot. This visualization hints that \method learns better feature manifolds in the few-shot classification setting.

\paragraph{{More few-shot settings \& Full data training.}} {Training on 32 shots, \methodvalfree yields 77.79\% average accuracy over 11 datasets and 10 seeds, better than lower-shot results (cf. \cref{tab:results_wo_validation_tab}). Using full data for training, \methodvalfree improves to 81.03\%
compared to 79.97\% for TaskRes. Of note, Tip-adapter-F trains $N_\text{data} {\times} D$ parameters, thus $1.3$B parameters for full ImageNet, which is not feasible. This also highlights the benefit of \method for which the number of trainable parameters is not a function of the dataset size.}
\paragraph{{Effect of temperature.}} {In all the experiments of the paper, we use the pretrained temperature value \mbox{temp$=1/\tau=100$} (cf. \cref{eq:proba_ik}). Here we ablate this choice and show in \cref{tab:temperature_effect} the performance of \methodvalfree for \mbox{temp$=50$} and \mbox{temp$=150$}. We observe that the performance is not highly affected, and that temp$=50$ even outperforms the pretrained value. Studying in depth the effect of this parameter is left for future research.} 

\begin{table}[t]
\newcommand{\varh}[1]{} 

\setlength{\fboxsep}{1pt}
\centering
\resizebox{0.55\linewidth}{!}{%
\setlength{\tabcolsep}{0.3em}
\begin{tabular}{l c c c c c}
\toprule
{temp} & $N=1$ & 2 & 4 & 8 & 16 \\
\midrule[0.7pt]
50 &  \best{64.47} & \best{67.37} & \best{70.25} & \best{73.13} & \best{75.72} \\
100 & \second{64.40} & \second{67.28} & \second{70.08} & \second{72.97} & \second{75.57}\\
150 & 64.15 & 67.03 & 69.85 & 72.70 & 75.24 \\

\bottomrule
\end{tabular}%
}%
\caption{{\textbf{Effect of temperature (temp).} We report classification accuracy (\%) of \methodvalfree averaged over 11 datasets and 10 seeds for different temperature values. LR=$10^{-4}$ and $\lambda=1/N$ for all datasets.}}
\label{tab:temperature_effect}
\end{table}

\section{\method training details}
\label{sec:add_training_details}
The text encoder is fully frozen during training of \method.
The templates are similar to previous works~\citep{zhang2022tip,huang2024lp++} for fair comparison, and are detailed in~\cref{tab:templates} for each dataset.

\begin{table}[h]
    \centering
    \resizebox{\linewidth}{!}{%
    \begin{tabular}{l c}
        \toprule
        \textbf{Dataset} & \textbf{Template} \\
        \midrule
        Caltech101 & ``a photo of a \{class\}.'' \\
        StanfordCars & ``a photo of a \{class\}.'' \\
        SUN397 & ``a photo of a \{class\}.'' \\
        DTD & ``\{class\} texture.'' \\
        Eurosat & ``a centered satellite photo of \{class\}.'' \\
        FGVCAircraft & ``a photo of a \{class\}, a type of aircraft.'' \\
        Food101 & ``a photo of \{class\}, a type of food.'' \\
        Flowers102 & ``a photo of a \{class\}, a type of flower.'' \\
        OxfordPets & ``a photo of a \{class\}, a type of pet.'' \\
        UCF101 & ``a photo of a person doing \{class\}.'' \\
        ImageNet & \multirow{5}{*}{\rule[0.5em]{1pt}{6em}\raisebox{3.2em}[0pt][0pt]{
        \parbox{8cm}{\centering%
        Ensemble of 7 templates:\\%
        \{``itap of a \{class\}.'', ``a bad photo of the \{class\}.'',\\%
        ``a origami \{\}.'', ``a photo of the large \{class\}.'',\\%
        ``a \{class\} in a video game.'', ``art of the \{class\}.'',\\%
        ``a photo of the small \{class\}.''\}}}}\\
        ImageNet-A & \\
        ImageNet-V2 & \\
        ImageNet-R & \\
        ImageNet-Sketch & \\
        
        \bottomrule
    \end{tabular}
    }
    \caption{\textbf{Dataset-specific templates.} Following the literature, all but ImageNet dataset and its variants use a single template.}
    \label{tab:templates}
\end{table}

For training, only the weight matrix $\mW_o$ in~\cref{eq:resnet_processing} is fine-tuned. Note that for ResNets, a bias term $\vb_o$ exists while for ViTs no bias is added in pretraining. We stress that fine-tuning also the bias term for ResNets does not change the results, as most of the parameters as concentrated in the weight matrix. In detail for ResNet-50, $\mW_o \in \mathbb{R}^{D \times D_o}$, where $D=1024$ and $D_o=2048$, this makes a total of $~\sim${2M} parameters, while $b_o \in \mathbb{R}^{D}$ has only $1024$ parameters.~\cref{tab:parameters} shows the number of trainable parameters in \method for different backbones.

\begin{table}[h]
    \centering
    \resizebox{0.7\linewidth}{!}{%
    \begin{tabular}{l c c}
        \toprule
        \textbf{Backbone} & $D \times D_o$ &\textbf{Parameters in $W_o$} \\
        \midrule
        ResNet-50 & $1024\times 2048$ & 2.097M\\
        ResNet-101 & $512\times 2048$ & 1.049M\\
        ViT-B/32 & $512 \times 768$ & 0.393M\\
        ViT-B/16 & $512 \times 768$ & 0.393M\\
        \bottomrule
    \end{tabular}
    }
    \caption{\textbf{Number of trainable parameters per backbone.} It is the number of elements in the projection matrix $W_o \in \mathbb{R}^{D \times D_o}$.}
    \label{tab:parameters}
\end{table}

\begin{table}[h]
\centering
\renewcommand{\arraystretch}{1.2}
\resizebox{0.8\linewidth}{!}{
\setlength{\tabcolsep}{0.5em}
\begin{tabular}{llccccc}
\toprule
 Dataset & \multicolumn{1}{c}{Method}  & $N=1$ & 2 & 4 & 8 & 16\\
\midrule
\multirow{10}{*}{\textit{ImageNet}} & \textcolor{gray}{CLIP (0-shot)} & \multicolumn{5}{c}{\textcolor{gray}{60.35}}\\
\cmidrule{2-7}
& CoOp~\citep{zhou2022learning} & $61.19$ & $61.58$  &  $62.22$  & $62.87$  &  $63.70$ \\
& PLOT~\citep{chen2023plot} & $60.46$ &  $60.73$ &  $61.79$  &  $62.48$ & $63.08$  \\
& KgCoOp~\citep{yao2023visual} & $60.90$ & $61.44$ & $62.00$ &$62.20$ & $62.43$ \\
& ProGrad~\citep{zhu2023prompt} & $\bm{61.58}$ & $\bm{62.14}$ & $\bm{62.59}$ & $63.04$ & $63.54$\\
\cmidrule{2-7}

& CLIP-Adapter~~\citep{gao2024clip} & $59.82 $ & $59.94$ & $59.97$ & $59.98$ & $ 61.31$ \\
& Tip-Adapter-F~\citep{zhang2022tip} & $60.59 $ & $61.42 $ & $62.12 $  & $63.41$  & $\bm{65.06}$ \\
& Tip-Adapter-F*~\citep{zhang2022tip} & $60.98 $ & $61.23 $ & $61.72 $ & $62.84$ & $64.03$ \\
\cmidrule{2-7}
& Standard LP~\citep{radford2021learning} & $22.21$  &  $31.96 $  &  $41.48 $  &  $49.49 $  & $56.04 $  \\
& LP++~\citep{huang2024lp++} & {$61.18$} &  {$61.56$} & {$62.55$} & {$\bm{63.76}$} & {$64.73$}\\
\cmidrule{2-7}
& \method & $61.28$ & $61.79$ & $62.38$ & $63.30$ & $64.31$ \\
\bottomrule

\multirow{10}{*}{\textit{SUN397}} & \textcolor{gray}{CLIP (0-shot)} & \multicolumn{5}{c}{\textcolor{gray}{58.85}}\\
\cmidrule{2-7}
& CoOp~\citep{zhou2022learning} &$61.79$ &  $63.32$ & $65.79$ & $67.89$  &  $70.15$ \\
& PLOT~\citep{chen2023plot} & $62.53$ & $63.87$ & $65.85$ & $67.83$ & $69.90$  \\
& KgCoOp~\citep{yao2023visual} & $62.91$  &  $64.38$ & $66.06$ & $66.66$ & $67.68$ \\
& ProGrad~\citep{zhu2023prompt} & $62.79$ & $64.12$ & $66.32$ & $68.33$ & $70.18$\\
\cmidrule{2-7}
& CLIP-Adapter~\citep{gao2024clip} & $60.78 $ & $61.79$ & $63.84$ & $66.26$ & $ 67.66$ \\
& Tip-Adapter-F~\citep{zhang2022tip} & $61.02$  & $62.15$  & $63.86$ & $67.25$ & $70.94$ \\
& Tip-Adapter-F*~\citep{zhang2022tip} & $62.58$  & $63.79$  & $65.49$ & $67.43$ & $69.25$ \\
\cmidrule{2-7}
& Standard LP~\citep{radford2021learning} & $32.56$  & $43.77$   &  $54.49$  &  $61.83$  &  $67.03 $  \\
& LP++~\citep{huang2024lp++} & {$62.47$} & {$64.65$} & {$67.28$} & {$\bm{69.34}$} & {$71.23$}\\
\cmidrule{2-7}
& \method & $\bm{63.44}$ & $\bm{65.16}$ & $\bm{67.39}$ & $69.31$ & $\bm{71.31}$ \\
\bottomrule

\multirow{10}{*}{\textit{DTD}} & \textcolor{gray}{CLIP (0-shot)} & \multicolumn{5}{c}{\textcolor{gray}{42.69}}\\
\cmidrule{2-7}
& CoOp~\citep{zhou2022learning} & $42.31$ &  $47.13$ &  $54.06$  & $59.21$  &  $63.67$ \\
& PLOT~\citep{chen2023plot} & $45.82$ & $51.32$  & $55.67$   &  $61.38$ &  $65.29$  \\
& KgCoOp~\citep{yao2023visual} & $45.46$ & $50.01$ & $53.37$ &  $58.38$  & $62.71$ \\
& ProGrad~\citep{zhu2023prompt} & $44.19$  & $ 50.41$  & $54.82$ & $60.31$ & $63.89$\\
\cmidrule{2-7}
& CLIP-Adapter~\citep{gao2024clip} & $43.49 $ & $44.49$ & $48.95$ & $57.52$ & $ 62.97$ \\
& Tip-Adapter-F~\citep{zhang2022tip} & $46.92$ & $48.50$ & $57.16$ & $62.38$ & $65.23$ \\
& Tip-Adapter-F*~\citep{zhang2022tip} & $47.68$ & $52.24$ & $56.09$ & $61.05$ & $65.04$ \\
\cmidrule{2-7}
& Standard LP~\citep{radford2021learning} & $29.63$ & $41.19$   & $51.72$   &  $58.78$  & $64.56$  \\
& LP++~\citep{huang2024lp++} & {$46.97$} & {$52.44$} & {$57.75$} & {$62.42$} & {$66.40$}\\
\cmidrule{2-7}
& \method & $\bm{50.21}$ & $\bm{54.75}$ & $\bm{59.30}$ & $\bm{64.19}$ & $\bm{68.02}$\\
\bottomrule

\multirow{10}{*}{\textit{Caltech101}} & \textcolor{gray}{CLIP (0-shot)} & \multicolumn{5}{c}{\textcolor{gray}{85.84}}\\
\cmidrule{2-7}
& CoOp~\citep{zhou2022learning} & $87.06$ & $89.14$  &  $90.00$  &  $91.00$ &  $91.77$ \\
& PLOT~\citep{chen2023plot} & $\bm{89.41}$ & $\bm{90.22}$  &  $90.69$  &  $91.55$ &  $92.17$  \\
& KgCoOp~\citep{yao2023visual} & $88.24$ & $88.85$  &  $89.89$ & $90.32$ & $90.93$ \\
& ProGrad~\citep{zhu2023prompt} & $88.34$  & $89.01$  & $90.13$  & $90.76$ & $91.67$\\
\cmidrule{2-7}
& CLIP-Adapter~\citep{gao2024clip} & $87.69 $ & $89.37$ & $90.21$ & $91.33$ & $ 92.10$ \\
& Tip-Adapter-F~\citep{zhang2022tip} & $87.35$ & $88.17$ & $89.49$ & $90.54$ & $92.10$ \\
& Tip-Adapter-F*~\citep{zhang2022tip} & $88.68$ & $89.36$ & $90.40$ & $91.62$ & $92.63$ \\
\cmidrule{2-7}
& Standard LP~\citep{radford2021learning} & $68.88$  &  $78.41$  &  $84.91$  &  $88.70$  &  $91.14$  \\
& LP++~\citep{huang2024lp++} & {$88.56$} & {$89.53$} & {$90.87$} & {$91.84$} & {$92.73$}\\
\cmidrule{2-7}
& \method & $89.25$ & $89.80$ & $\bm{91.47}$ & $\bm{92.37}$ & $\bm{93.44}$ \\
\bottomrule

\multirow{10}{*}{\textit{UCF101}} & \textcolor{gray}{CLIP (0-shot)} & \multicolumn{5}{c}{\textcolor{gray}{61.80}}\\
\cmidrule{2-7}
& CoOp~\citep{zhou2022learning} & $62.80$ &  $65.62$ &  $68.69$ & $72.57$ & $76.39$ \\
& PLOT~\citep{chen2023plot} &  $63.22$ & $66.49$ & $70.12$ & $74.63$ & $77.39$ \\
& KgCoOp~\citep{yao2023visual} & $64.37$ & $64.91$ & $68.41$ &  $69.86$ & $71.73$ \\
& ProGrad~\citep{zhu2023prompt} & $65.13$ & $66.57$ & $69.80$ & $73.01$ & $75.76$ \\
\cmidrule{2-7}
& CLIP-Adapter~\citep{gao2024clip} & $64.25$ & $66.68$ & $69.77$ & $73.90$ & $77.26$ \\
& Tip-Adapter-F~\citep{zhang2022tip} & $64.28$ & $65.48$ & $67.61$ & $72.05$ & $77.30$ \\
& Tip-Adapter-F*~\citep{zhang2022tip} & $65.50$ & $68.55$ & $70.55$ & $74.25$ & $76.83$ \\
\cmidrule{2-7}
& Standard LP~\citep{radford2021learning} & $40.80$ & $51.71$ & $61.64$ & $68.47$ & $73.38$ \\
& LP++~\citep{huang2024lp++} & $65.41$ & $69.20$ & $71.68$ & $74.86$ & $77.46$ \\
\cmidrule{2-7}
& \method & $\bm{67.88}$ & $\bm{70.07}$ & $\bm{73.51}$ & $\bm{77.06}$ & $\bm{79.79}$\\
\bottomrule

\multirow{10}{*}{\textit{Flowers102}} & \textcolor{gray}{CLIP (0-shot)} & \multicolumn{5}{c}{\textcolor{gray}{65.98}}\\
\cmidrule{2-7}
& CoOp~\citep{zhou2022learning} &  $69.00$ & $78.47$  & $85.34$   &  $91.68$ & $94.47$ \\ 
& PLOT~\citep{chen2023plot} & $71.09$ &  $81.22$ &  $87.61$  &  $92.60$ &  $\bm{95.18}$ \\
& KgCoOp~\citep{yao2023visual} & $68.73$ & $69.63$ &  $76.51$ & $80.71$ & $84.48$ \\
& ProGrad~\citep{zhu2023prompt} & $72.16$ & $79.55$ & $84.56$ & $91.73$ &  $94.10$ \\
\cmidrule{2-7}
& CLIP-Adapter~\citep{gao2024clip} & $66.86 $ & $69.71$ & $77.42$ & $87.20$ & $ 91.16$ \\
& Tip-Adapter-F~\citep{zhang2022tip} & $67.73$ & $68.18$ & $71.17$ & $84.11$ & $93.02$ \\
& Tip-Adapter-F*~\citep{zhang2022tip} & $\bm{78.46}$ & $\bm{85.14}$ & $88.53$ & $92.33$ & $94.26$ \\
\cmidrule{2-7}
& Standard LP~\citep{radford2021learning} &  $56.98$  &  $73.40$  &  $84.38$  &  $91.81$  & $95.05$ \\
& LP++~\citep{huang2024lp++} & $78.21$ & $84.69$ & $\bm{89.56}$ & $92.61$ &  $94.26$ \\
\cmidrule{2-7}
& \method & $75.33$ & $81.95$ & $88.34$ & $\bm{92.68}$ & $94.92$ \\
\bottomrule
\end{tabular}
}
\caption{\textbf{Comparison to state-of-the-art methods}. Average classification accuracy (\%) and standard deviation over 10 tasks for 11 benchmarks. Best values are highlighted in bold.}
\label{tab:per-dataset-perf-1}
\end{table}

\begin{table}[t]
\centering
\renewcommand{\arraystretch}{1.2}
\resizebox{0.9\linewidth}{!}{
\setlength{\tabcolsep}{0.5em}
\begin{tabular}{llccccc}
\toprule
 Dataset & \multicolumn{1}{c}{Method}  & $N=1$ & 2 & 4 & 8 & 16\\
\midrule
\multirow{10}{*}{\textit{StanfordCars}} & \textcolor{gray}{CLIP (0-shot)} & \multicolumn{5}{c}{\textcolor{gray}{55.78}}\\
\cmidrule{2-7}
& CoOp~\citep{zhou2022learning} & $57.00$& $58.96$ & $62.81$ & $68.40$ & $72.87$ \\
& PLOT~\citep{chen2023plot} & $57.47$& $59.89$& $63.49$&  $68.75$ &  $73.86$  \\
& KgCoOp~\citep{yao2023visual} & $57.19$  & $58.94$ & $59.85$ & $61.42$ & $62.99$ \\
& ProGrad~\citep{zhu2023prompt} & $58.63$ & $61.23$ & $65.02$ & $69.43$ & $72.76$ \\
\cmidrule{2-7}

& CLIP-Adapter~~\citep{gao2024clip} & $56.67 $ & $57.94$ & $61.13$ & $65.43$ & $ 70.24$ \\
& Tip-Adapter-F~\citep{zhang2022tip} & $57.24$ & $58.12$ & $59.34$ & $64.25$ & $71.38$ \\
& Tip-Adapter-F*~\citep{zhang2022tip} & $57.85$ & $60.55$ & $64.22$ & $68.75$ & $74.19$ \\
\cmidrule{2-7}
& Standard LP~\citep{radford2021learning} & $22.94$  &  $35.48$  &  $47.49$  &  $59.34$  & $69.11$ \\
& LP++~\citep{huang2024lp++} & $57.20$ & $59.95$ & $63.44$ & $67.81$ & $72.33$ \\
\cmidrule{2-7}
& \method & $\bm{58.72}$ & $\bm{61.71}$ & $\bm{65.68}$ & $\bm{70.64}$ & $\bm{75.64}$ \\
\midrule[0.7pt]

\multirow{10}{*}{\textit{FGVCAircraft}} & \textcolor{gray}{CLIP (0-shot)} & \multicolumn{5}{c}{\textcolor{gray}{17.07}}\\
\cmidrule{2-7}
& CoOp~\citep{zhou2022learning} & $12.50$& $17.59$ & $21.27$& $26.85$& $31.20$ \\
& PLOT~\citep{chen2023plot} & $17.75$ & $19.55$ & $22.26$& $26.70$  & $32.09$ \\
& KgCoOp~\citep{yao2023visual} & $ 18.61$  &  $18.93$ & $21.16$ & $22.80$  & $24.10$  \\
& ProGrad~\citep{zhu2023prompt} & $18.41$  & $20.51$ & $23.65$ & $26.98$ & $30.47$ \\
\cmidrule{2-7}
& CLIP-Adapter~\citep{gao2024clip} & $18.56 $ & $19.18$ & $21.00$ & $23.76$ & $ 33.37$ \\
& Tip-Adapter-F~\citep{zhang2022tip} & $18.23$ & $19.12$ & $20.55$ & $23.60$ & $30.37$ \\
& Tip-Adapter-F*~\citep{zhang2022tip} & $19.08$ & $20.79$ & $23.99$ & $30.58$ & $36.16$ \\
\cmidrule{2-7}
& Standard LP~\citep{radford2021learning} & $12.66$ &  $16.92$  &  $21.11$  &  $26.53$  &  $32.42$  \\
& LP++~\citep{huang2024lp++} & $19.69$ & $21.58$ & $24.22$ & $27.73$ & $31.73$\\
\cmidrule{2-7}
& \method & $\bm{19.74}$ & $\bm{22.68}$ & $\bm{27.08}$ & $\bm{33.20}$ & $\bm{39.90}$ \\
\midrule[0.7pt]

\multirow{10}{*}{\textit{EuroSAT}} & \textcolor{gray}{CLIP (0-shot)} & \multicolumn{5}{c}{\textcolor{gray}{36.22}}\\
\cmidrule{2-7}
& CoOp~\citep{zhou2022learning} & $40.36$& $56.15$& $66.13$&  $77.02$ & $82.59$ \\
& PLOT~\citep{chen2023plot} & $44.22$& $64.19$ & $69.37$&  $78.84$ & $81.76$ \\
& KgCoOp~\citep{yao2023visual} & $43.86$  &  $ 52.92$  &  $ 59.51$  &  $63.23$   &  $64.04$ \\
& ProGrad~\citep{zhu2023prompt} & $49.37$ & $65.22$ & $69.57$  & $78.44$ & $82.17$ \\
\cmidrule{2-7}
& CLIP-Adapter~\citep{gao2024clip} & $43.00 $ & $48.60$ & $59.15$ & $69.92$ & $ 75.38$ \\
& Tip-Adapter-F~\citep{zhang2022tip} & $47.63$ & $57.62$ & $69.30$ & $75.22$ & $78.59$ \\
& Tip-Adapter-F*~\citep{zhang2022tip} & $49.27$ & $65.66$ & $70.72$ & $74.66$ & $78.73$ \\
\cmidrule{2-7}
& Standard LP~\citep{radford2021learning} & $48.29$  & $56.81$   &  $64.99$  & $74.56$   &  $80.29$ \\
& LP++~\citep{huang2024lp++} & $57.23$ & $61.65$ & $68.67$ & $75.86$ &  $80.53$\\
\cmidrule{2-7}
& \method & $\bm{57.95}$ & $\bm{70.03}$ & $\bm{76.48}$ & $\bm{81.81}$ & $\bm{85.81}$\\
\midrule[0.7pt]

\multirow{10}{*}{\textit{OxfordPets}} & \textcolor{gray}{CLIP (0-shot)} & \multicolumn{5}{c}{\textcolor{gray}{85.75}}\\
\cmidrule{2-7}
& CoOp~\citep{zhou2022learning} & $86.27$ &  $86.33$ & $85.34$   & $87.85$  & $88.68$ \\
& PLOT~\citep{chen2023plot} & $87.15$ & $87.23$  &  $88.03$  & $88.38$  &  $88.23$ \\
& KgCoOp~\citep{yao2023visual} & $87.51$  & $87.51$ & $88.04$ & $88.59$ &  $89.28$ \\
& ProGrad~\citep{zhu2023prompt} & $\bm{88.34}$  & $\bm{87.88}$ & $\bm{88.59}$ & $\bm{88.87}$ & $\bm{89.39}$ \\
\cmidrule{2-7}
& CLIP-Adapter~\citep{gao2024clip} & $85.46 $ & $86.37$ & $87.21$ & $87.95$ & $ 88.33$ \\
& Tip-Adapter-F~\citep{zhang2022tip} & $85.70$ & $86.05$ & $86.40$ & $87.66$ & $89.08$ \\
& Tip-Adapter-F*~\citep{zhang2022tip} & $86.05$ & $86.49$ & $87.19$ & $87.89$ & $88.26$ \\
\cmidrule{2-7}
& Standard LP~\citep{radford2021learning} & $30.62$  &  $42.64$  &  $55.60$  &  $67.32$  & $76.23$ \\
& LP++~\citep{huang2024lp++} & $84.24$ & $85.74$ & $86.94$ & $87.71$ & $88.38$\\
\cmidrule{2-7}
& \method & $85.46$ & $86.17$ & $87.05$ & $88.15$ & $89.17$ \\
\bottomrule

\multirow{10}{*}{\textit{Food101}} & \textcolor{gray}{CLIP (0-shot)} & \multicolumn{5}{c}{\textcolor{gray}{77.35}}\\
\cmidrule{2-7}
& CoOp~\citep{zhou2022learning} & $75.58$  &  $77.49$ & $77.93$  &  $78.92$ & $79.21$ \\
& PLOT~\citep{chen2023plot} &  $77.46$ & $77.72$ & $78.23$ & $78.40$ & $78.86$ \\
& KgCoOp~\citep{yao2023visual} & $77.20$  & $\bm{78.04}$  & $77.97$ & $78.39$  &  $78.73$ \\
& ProGrad~\citep{zhu2023prompt} & $\bm{78.36}$  & $78.01$ & $\bm{78.38}$ & $\bm{79.11}$ & $\bm{79.51}$ \\
\cmidrule{2-7}
& CLIP-Adapter~\citep{gao2024clip} & $76.93 $ & $77.22$ & $77.64$ & $77.97$ & $ 78.45$ \\
& Tip-Adapter-F~\citep{zhang2022tip} &  $77.53$  & $77.53$  & $77.82$ & $78.26$ & $78.99$ \\
& Tip-Adapter-F*~\citep{zhang2022tip} & $77.58$ & $77.36 $ & $77.78 $ & $78.17$ & $78.72$ \\
\cmidrule{2-7}
& Standard LP~\citep{radford2021learning} & $31.59$  &  $44.60$  & $56.13$   &  $64.45$  &  $70.97$  \\
& LP++~\citep{huang2024lp++} & $76.61$ & $77.22$ & $77.79$ & $78.53$ & $78.88$ \\
\cmidrule{2-7}
& \method & $77.06$ & $77.61$ & $77.74$ & $78.37$ & $79.21$ \\
\bottomrule

\end{tabular}
}
\caption{\textbf{Comparison to state-of-the-art methods} (Continued). Average classification accuracy (\%) and standard deviation over 10 tasks for 11 benchmarks. Best values are highlighted in bold.}
\label{tab:tab:per-dataset-perf-2}    
\end{table}

\section{Preliminaries}
\label{sec:preliminaries_supp}

\subsection{Zero-shot classification}
\label{sec:__}
We denote $\displaystyle f$ and $\displaystyle g$ the vision and text encoders of CLIP, respectively. During pretraining, CLIP learns a joint embedding space that pulls corresponding image-text representations closer together and pushes away dissimilar ones. At inference, given an image $\displaystyle \tI$, one only needs the names of $K$ candidate classes to perform \textit{zero-shot classification}: 
\begin{equation}
\hat{k} = \text{argmax}_k \displaystyle \vv^\intercal\vt_k,
\end{equation}
where $\vv = \displaystyle \frac{f(\tI; \theta_f)}{\|f(\tI; \theta_f)\|_2}$, $\vt_k = \displaystyle \frac{g(\mT_k; \theta_g)}{\|g(\mT_k; \theta_g)\|_2}$; $\displaystyle \theta_f$ and $\displaystyle \theta_g$ are the frozen parameters of $\displaystyle f$ and $\displaystyle g$, respectively; $\displaystyle \mT_k$ is a text prompt describing the class $k$, \textit{e.g.}, ``a photo of $\{\text{class}_k\}$''.

\subsection{Few-shot classification}
Given a set of $N$ labeled samples from each of the $K$ classes, research has been carried out to efficiently adapt CLIP using this set. All existing research in this direction can be gathered in three main avenues (see \cref{fig:teaser_figure}).

\medskip\noindent\textbf{Prompt Tuning.} It parameterizes the prompt template, \textit{i.e.}, $\displaystyle \mT_k = [\vw]_1[\vw]_2...[\vw]_M[\text{class}_k]$, where $[\vw]_1$, $[\vw]_2,\ldots,$ and $[\vw]_M$ are learned while keeping $\displaystyle f$ and $\displaystyle g$ frozen. Prompt tuning adapts CLIP ``indirectly'' on the classifier side, \textit{i.e.}, the text embeddings are derived from the learned prompts.

\medskip\noindent\textbf{Adapters.} They learn a multi-layer perceptron~(MLP) $\displaystyle h_{\theta}$ with a residual connection $\displaystyle \alpha$ on top of the frozen visual features $\vv$, \textit{i.e.}, $\displaystyle \vv := \alpha \vv + (1-\alpha) h_{\theta}(\vv)$, or on top of the frozen text features $\vt$, \textit{i.e.}, $\displaystyle \vt := \alpha \vt + (1-\alpha) h_{\theta}(\vt)$, or both.

\medskip\noindent\textbf{Linear probing.} It trains a linear classifier \mbox{$\displaystyle \mW \in \mathbb{R}^{K \times D}$} on top of the frozen visual features, $D$ being the embedding space dimension. Matrix $\mW$ can be initialized with text embeddings $\vt_k$. Since the classifier is directly tuned, linear probing restricts CLIP to $K$ classes after adaptation and cannot be applied in open-class setting.

\subsection{CLIP architecture}
\label{sec:__}

CLIP adopts a transformer architecture~\citep{vaswani2017attention} for the text encoder, but the vision encoder may be either a ResNet~\citep{he2016deep} or a Vision Transformer (ViT)~\citep{dosovitskiy2021an}.
We detail both architectures below and later elaborate on our unified method applicable to both architectures regardless of their intrinsic differences.

\medskip\noindent\textbf{ResNet.} CLIP replaces the global average pooling layer in ResNet with an attention pooling layer. The output of the multi-head attention layer is then projected to the shared latent space using a linear layer. Thus, $\displaystyle f$ can be written as $\displaystyle f = f_2 \circ f_1$, where $\displaystyle f_1$ represents all the layers up to the attention pooling (included), and $\displaystyle f_2$ represents the linear projection head. Given an image $\displaystyle \tI$:
\begin{equation}
    \label{eq:__}
    \vx_o = {f_1}(\tI) , \quad \vv = f_2(\vx_o) = \mW_o \vx_o + \vb_o,
\end{equation}
with $\vx_o \in \mathbb{R}^{D_o}$ the output of the attention pooling layer, $\mW_o \in \mathbb{R}^{D \times D_o}$ the projection matrix and $\vb_o$ a bias term.

\medskip\noindent\textbf{ViT.} The transformer encoder consists of multiple residual attention blocks. Each block has two main components: a multi-head self-attention and a feed-forward neural network (MLP), with residual connections.
The output of the last residual attention block is projected to the latent space using a trainable matrix. Thus, $\displaystyle f$ can be written as \mbox{$\displaystyle f = f_2 \circ f_1$}, where $\displaystyle f_1$ represents all the layers up to the last residual attention block (included), and $\displaystyle f_2$ represents the projection matrix. Given an image $\displaystyle \tI$:
\begin{equation}
\label{eq:___}
    \vx_o = {f_1}(\tI) , \quad \vv = {f_2}(\vx_o) = \mW_o \vx_o,
\end{equation}
where no bias term is included, unlike~\cref{eq:__}.

Similarly on the text side, the embeddings are projected into the shared latent space using a linear layer.

\end{document}